\documentclass[11pt]{article}

\PassOptionsToPackage{hyperfootnotes=false}{hyperref}
\usepackage[preprint]{acl}

\usepackage{times}
\usepackage{latexsym}
\usepackage[T1]{fontenc}
\usepackage[utf8]{inputenc}
\usepackage{microtype}
\usepackage{courier}

\usepackage{amsmath,amssymb,amsfonts}
\usepackage{float}
\floatstyle{ruled}
\newfloat{algorithm}{tbp}{loa}
\floatname{algorithm}{Algorithm}
\usepackage{graphicx}
\usepackage{booktabs}
\usepackage{multirow}
\usepackage{xcolor}
\usepackage{enumitem}
\usepackage[capitalize]{cleveref}
\crefname{algorithm}{Algorithm}{Algorithms}
\Crefname{algorithm}{Algorithm}{Algorithms}

\graphicspath{{figures/}}

\newcommand{\ours}{NavTree}
\newcommand{\pp}{\,\textrm{pp}}
\newcommand{\eg}{\emph{e.g.}}

\title{Tree Navigation Without LLM Summaries:\\
A Matched-Cost Study of Hierarchical Retrieval for Long-Document QA}

\author{Priyank Jayraj\thanks{Corresponding author.} \quad Poonam Goyal \quad Navneet Goyal \\
  Birla Institute of Technology and Science, Pilani \\
  Pilani, Rajasthan, India \\
  {\small\texttt{priyank.jayraj@pilani.bits-pilani.ac.in}} \\
  {\small\texttt{poonam@pilani.bits-pilani.ac.in}\quad\texttt{goel@pilani.bits-pilani.ac.in}}}

\hypersetup{
  pdftitle={Tree Navigation Without LLM Summaries: A Matched-Cost Study of Hierarchical Retrieval for Long-Document QA},
  pdfauthor={Priyank Jayraj, Poonam Goyal, Navneet Goyal}
}

\begin{document}
\maketitle

\begin{abstract}
Retrieval-augmented generation grounds language models in external context, but for long documents flat top-$k$ retrieval can cluster on a single region and miss complementary evidence. \textsc{RAPTOR}-style \emph{summary trees} address this by recursively clustering chunks and using a language model to summarize each cluster at indexing time, then ranking summary nodes alongside raw chunks at query time. We show the main benefit of summary trees in long-document QA can come from \emph{navigation} rather than the generated summary content. We introduce \ours{}, a leaves-only retriever that builds a deterministic balanced segment tree over chunks (zero language-model calls at indexing) and uses the tree purely as a navigation scaffold: a hybrid lexical-and-dense frontier walk, anchored on top retrieved leaves, descends from the root and emits only leaf chunks to the reader. On a matched-cost evaluation against flat retrievers and an extractive re-implementation of \textsc{RAPTOR}, \ours{} is the strongest matched-cost hierarchical retriever in our evaluated grid and ties the strongest flat baseline. On long-document multi-hop QA, it is the only hierarchical method that significantly beats BM25 on a class-vs-class basis, corroborated by a reader-free retrieval-recall check. A matched-reader replication of the published abstractive \textsc{RAPTOR} variant, given strong cluster summaries, still loses to \ours{} at every multi-chunk budget, at zero indexing cost. The ranking carries across stronger and open-weight readers, a stronger encoder, and a full factorial that isolates leaves-only emission as the structural lever.
\end{abstract}

\section{Introduction}
\label{sec:intro}

\noindent Retrieval-augmented generation (RAG) grounds large language models (LLMs) in external context~\cite{lewis2020rag,izacard2022atlas,ram2023incontext}. Most deployed systems use \emph{flat retrieval}: chunk documents into fixed-size segments, embed each independently with a dense encoder (\textsc{SBERT}~\cite{reimers2019sbert}, \textsc{DPR}~\cite{karpukhin2020dpr}), and return the top-$k$ by similarity. For questions that span multiple parts of a long document, flat retrieval can cluster on a single region of the document and miss complementary evidence.

\noindent\textbf{Hierarchical alternative.} \textsc{RAPTOR}~\cite{sarthi2024raptor} addresses this by recursively clustering chunks (GMM with UMAP and BIC), \emph{summarizing} each cluster with an LLM (\eg{}, GPT-3.5) at indexing time, and embedding the summaries to build a tree from leaves to root. At query time its \emph{collapsed tree} ranks every node, leaf or summary, by cosine similarity to the query and greedily packs the top-$k$ under a token budget; coupled with GPT-4 as the reader, \textsc{RAPTOR} reported $82.6\%$ on the QuALITY dev set~\cite{pang2022quality}, a state of the art at the time of publication. Related tree-style retrievers either build trees by summarizing adjacent chunks~\cite{liu2022llamaindex}, prepend LLM-derived structural titles~\cite{tao2025treerag}, combine document- and passage-level retrieval~\cite{liu2021dhr,jin2025longrefiner}, or exploit Rhetorical Structure Theory~\cite{chen2025disretrieval}; all of them include summary nodes in the retrieved context.

\noindent\textbf{An overlooked separation.} This design conflates two distinct roles of the tree: the tree as a \emph{navigation structure} (which subtree contains the answer?) and the tree as a \emph{content provider} (what summary text should we feed the reader?). The first role requires only the tree topology and node embeddings; the second requires high-quality summaries, which in \textsc{RAPTOR} demand one LLM call per internal node at indexing time and scale linearly with corpus size. We ask whether the second role is needed at all.

\begin{figure*}[t]
    \centering
    \includegraphics[width=\textwidth]{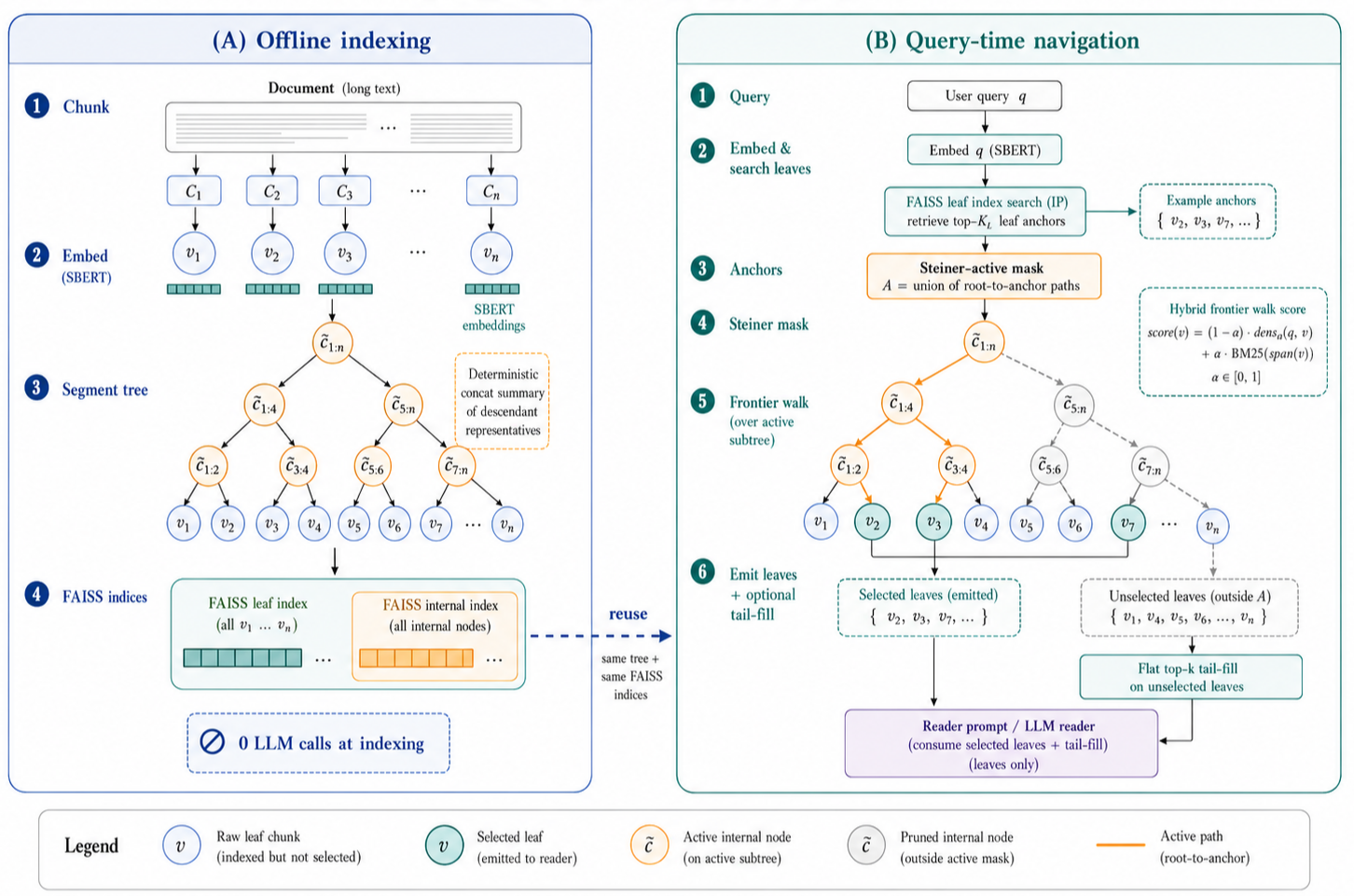}
    \caption{\textbf{Architecture of \ours{}.}
    \textbf{(A) Offline indexing.} NavTree splits a document into sentence-aware chunks, embeds them as leaf nodes $\{v_1,\ldots,v_n\}$, and builds a deterministic balanced segment tree. Internal nodes $\bar{c}_{i:j}$ store conservative concatenations of representative descendant sentences, but no LLM-generated summaries are produced. Leaf and internal-node embeddings are stored in separate FAISS flat inner-product indices, yielding zero LLM calls at indexing time.
    \textbf{(B) Query-time navigation.} The query embedding $\mathbf{q}$ retrieves top-$K_\ell$ leaf anchors, whose root-to-leaf paths define the Steiner-active mask $\mathcal{A}$. A hybrid dense--BM25 frontier walk descends through this active subtree. Internal nodes are routing-only: they are split but never emitted. The reader receives only selected raw leaf chunks, optionally complemented by a flat top-$k$ tail-fill over unselected leaves. The dashed connector denotes reuse of the same offline tree and FAISS indices.}
    \label{fig:architecture}
\end{figure*}

\noindent\textbf{Approach.} We introduce \ours{}, a leaves-only navigated retriever (\cref{fig:architecture}). \ours{} builds a balanced binary segment tree over chunks; each internal node holds a deterministic concatenation of representative sentences from its descendants (\emph{no LLM calls at indexing}). Top-$K_\ell$ leaf anchors retrieved via FAISS define a \emph{Steiner-active mask} that prunes the tree to the union of root-to-anchor paths. A hybrid (BM25\,+\,dense) frontier walk descends to relevant leaves; only leaf chunks are emitted to the reader; remaining budget is filled with a global flat top-$k$ tail. Internal nodes participate only in routing.

\noindent\textbf{Comparison protocol.} Our \emph{primary} comparison is against a matched-cost re-implementation of \textsc{RAPTOR} that replaces GPT-3.5 cluster summaries with deterministic extractive ones (\textit{RAPTOR-Extractive}, \cref{sec:setup}). This isolates the retrieval algorithm from the LLM-summary quality. We do not attempt a cross-paper head-to-head against the original abstractive \textsc{RAPTOR}+GPT-4 number, because the original GPT-4 checkpoint is no longer callable and any modern OpenAI reader confounds the comparison.

\noindent\textbf{Contributions.} \ours{} shows that the main benefit of summary trees in long-document QA can come from \emph{navigation} rather than generated summary content.

\textbf{Strongest matched-cost hierarchical retriever in our evaluated grid.} \ours{} leads the hierarchical class in 6 of 8 headline cells (QuALITY + QASPER, 4 budgets each; \cref{tab:main}), significantly beats extractive RAPTOR on QuALITY@1024/@2000 ($+4.4$/$+5.4$ absolute, $p<10^{-7}$, $N{=}2086$), and \emph{ties} the BM25\,+\,SBERT hybrid, the strongest flat baseline, at every QuALITY budget ($p>0.1$ paired), reaching the flat-retrieval ceiling without LLM-summary indexing cost.

\textbf{Class-vs-class win on multi-hop QA.} Across HotpotQA, 2WikiMQA, MuSiQue, and full-book NarrativeQA ($N{=}200$ each, $5$--$23$k-token contexts; \cref{tab:multihop}), \ours{} is the only hierarchical method that significantly beats a flat method ($+0.077$/$+0.069$ F1 over BM25 on 2WikiMQA@1024/@2000, $p<10^{-3}$/$p<0.005$). Reader-free retrieval recall (\cref{sec:appendix_recall_multihop}) corroborates: highest answer-presence@$k$ on 2WikiMQA@2000 ($94.9\%$) and MuSiQue@2000 ($68.5\%$).

\textbf{Mechanism: a token-packing effect.} A direct reader-side head-to-head ($N{=}115$) shows summaries actually beat leaves \emph{per-node} ($55.7\%$ vs.\ $50.4\%$ on QuALITY); \ours{}'s wins emerge only at multi-chunk budgets, where summaries crowd out distinct-subtree leaves the reader could have used (\cref{sec:analysis_cosine}; \cref{tab:summary_vs_leaf} in App.).

\textbf{Method and cost.} A leaves-only navigated retriever (\cref{sec:method}, Alg.~\ref{alg:nav}) combining a deterministic balanced segment tree, Steiner-active pruning, hybrid (BM25\,+\,dense) frontier scoring, and a flat top-$k$ tail-fill, with \emph{zero} LLM calls at indexing. Abstractive RAPTOR by contrast requires $682$ LLM calls on a $60$-doc slice (\$$0.14$ at \texttt{gpt-4o-mini}, ${\sim}\$4$ at \texttt{gpt-4o}), scaling to $\$60$k$+$ on million-doc corpora (\cref{tab:index_cost}).

\textbf{Robustness across four axes.} The matched-cost ranking carries (a)~under \texttt{gpt-4o} (\cref{tab:sota_quality}); (b)~under \emph{two} open-weight readers via local Ollama, with $4/4$ reader-dataset cells leading or tied and all $12$ paired comparisons directionally consistent (\cref{sec:appendix_openweight}); (c)~under a stronger encoder (BGE-large; \cref{sec:appendix_bge}); and (d)~under a full $2^4$ factorial that isolates the leaves-only emission rule, not the walk algorithm, as the structural lever and CE rerank as the dominant retrieval-side lever (\cref{sec:appendix_factorial}). Hyper-parameter defaults sit on a flat plateau within $1.5${\pp} of all sweep neighbors (\cref{fig:sweep}; \cref{tab:qasper_sweep} in App.).

\section{Related Work}
\label{sec:related}

\paragraph{Hierarchical retrieval.}
The hierarchical retrievers most related to \ours{} (\textsc{RAPTOR} and the family of tree-style retrievers it inspired) are described in \cref{sec:intro}. They differ from \ours{} in that they all include LLM- or rule-derived summary nodes in the retrieved context. We show that for matched-cost retrieval the tree topology alone suffices; the summaries are an unforced indexing expense.

\paragraph{Other RAG, budgets, and reranking.}
Foundational RAG~\cite{lewis2020rag} has been refined along retriever~\cite{karpukhin2020dpr,khattab2020colbert,santhanam2022colbertv2,gao2022hyde}, reader~\cite{izacard2022atlas,shi2023replug}, and orchestration axes~\cite{asai2024selfrag,yan2024crag,jiang2023flare,jeong2024adaptiverag}. Recent methods reorganize the retrieval unit (\textsc{HippoRAG}~\cite{gutierrez2024hipporag}, \textsc{GraphRAG}~\cite{edge2024graphrag}, \textsc{LongRAG}~\cite{jiang2024longrag}) or trigger retrieval dynamically~\cite{su2024dragin,kim2024toc}; these are largely orthogonal to our leaves-only navigation question. Budget-aware approaches range from redundancy-aware scoring~\cite{peng2025adagres,wang2024corag} to post-hoc compression~\cite{jiang2023llmlingua,xu2024recomp}; \ours{} narrows the search space before content is selected. Cross-encoder reranking~\cite{nogueira2019monobert} is a standard two-stage pipeline; we use it only as an optional tight-budget add-on. \citet{liu2023lostmiddle} and \citet{yoran2024making} discuss how readers exploit (or fail to exploit) retrieved context; our leaves-only design contributes by avoiding paraphrased intermediate text.

\section{Method: \ours{}}
\label{sec:method}

\ours{} has three components: (i) a balanced segment tree built deterministically from raw chunks, (ii) Steiner-active anchoring on FAISS top-$k$ leaves, and (iii) a hybrid frontier walk that returns only leaves, with flat top-$k$ tail-fill.

\subsection{Segment tree construction}

Given a document chunked into $n$ segments $\{c_1,\dots,c_n\}$ (sentence-boundary splitting, $\approx 100$ tokens each), \ours{} builds a balanced binary segment tree. Each leaf $v_i$ stores chunk $c_i$, its normalized SBERT embedding $\mathbf{e}_i = \text{normalize}(\text{enc}(c_i))$, and its token cost $w_i$. Each internal node $v$ stores a deterministic, query-independent ``conservative'' payload (concatenation of representative sentences from its descendants) and the normalized embedding of that payload. The tree has $2n-1$ nodes and depth $\lceil\log_2 n\rceil$.

Two FAISS~\cite{johnson2021faiss} flat inner-product indices are built: one over the $n$ leaf embeddings, one over the $n-1$ internal embeddings. Tree construction is $O(n\log n)$ embedding calls (the same asymptotic cost as flat retrieval) and has \textbf{zero LLM calls}.

\subsection{Steiner-active anchoring}
\label{sec:steiner}

Given query embedding $\mathbf{q}$, retrieve the top-$K_\ell$ leaves via FAISS. (Optionally, in a ``classic'' mode that we do \emph{not} use as default, also retrieve top-$K_i$ internal nodes.) These nodes serve as \emph{anchors}. The \textbf{Steiner-active mask} is the union of root-to-anchor paths:
\begin{equation}
\mathcal{A} = \bigcup_{a \in \text{anchors}} \text{path}(\text{root}, a).
\end{equation}
Nodes outside $\mathcal{A}$ are pruned from navigation. The mask guarantees that every top-$k$ anchored leaf remains \emph{reachable} during the frontier walk: navigation cannot dead-end before exploring the right subtree.

\subsection{Hybrid frontier walk + flat fill}

\Cref{alg:nav} describes the retrieval procedure.

\begin{algorithm}[t]
\caption{\ours{}: navigated leaves-only retrieval with flat fill}\label{alg:nav}
\small
\noindent\textbf{Require:} segment tree $\mathcal{T}$, query embedding $\mathbf{q}$, budget $B$, anchor count $K_\ell$, hybrid weight $\alpha\in[0,1]$. Define the min--max normalizer over the current frontier $\mathcal{F}$ as
\[
\widetilde{x}_v = \frac{x_v - \min_{u\in\mathcal{F}} x_u}{\max_{u\in\mathcal{F}} x_u - \min_{u\in\mathcal{F}} x_u + \epsilon}
\]
for $x \in \{\cos(\mathbf{q},\cdot),\,\beta_{\mathrm{span}(\cdot)}\}$, re-normalized on every loop pass; the flat tail-fill in lines~\ref{algline:tail_start}--\ref{algline:tail_end} ranks against the global leaf set using percentile-rank versions of the same two scores.

\noindent\textbf{Ensure:} leaf set $S$ with $\sum_{v\in S} w_v \leq B$
\begin{enumerate}[label=\arabic*:,ref=\arabic*,leftmargin=1.8em,labelsep=.4em,itemsep=0pt,parsep=0pt,topsep=3pt]
\item $\mathrm{Anchors} \gets \mathrm{FAISS\text{-}topK}(\mathbf{q}, K_\ell)$ on leaves
\item $\mathcal{A} \gets \bigcup_{a\in\mathrm{Anchors}} \mathrm{path}(\mathrm{root}, a)$
\item $S \gets \emptyset$; $b \gets B$; $\mathcal{F} \gets \{\mathrm{root}\}$
\item \textbf{while} $\mathcal{F} \neq \emptyset$ and $b > 0$ \textbf{do}
\item \label{algline:score}\hspace*{.8em}$\begin{aligned}[t]\mathrm{score}(v) &\gets (1{-}\alpha)\widetilde{\cos}(\mathbf{q},\mathbf{e}_v)\\&\quad + \alpha\,\widetilde{\beta}_{\mathrm{span}(v)}\end{aligned}$\par
\hspace*{.8em}{\footnotesize$\triangleright$ $\widetilde{\cdot}$ = min--max over $\mathcal{F}$}
\item \hspace*{.8em}$v^* \gets \arg\max_{v\in\mathcal{F}} \mathrm{score}(v)$
\item \hspace*{.8em}\textbf{if} $v^*$ is leaf \textbf{then}
\item \hspace*{1.6em}\textbf{if} $w_{v^*}\leq b$ \textbf{then}
\item \hspace*{2.4em}$S \gets S \cup \{v^*\}$; $b \gets b - w_{v^*}$
\item \hspace*{1.6em}\textbf{end if}\par
\hspace*{1.6em}{\footnotesize$\triangleright$ Oversize leaves are picked up by flat tail-fill below.}
\item \hspace*{.8em}\textbf{else if} $v^*$ is internal \textbf{then}
\item \label{algline:split}\hspace*{1.6em}$\mathcal{F} \gets \mathcal{F} \cup \{c \in \mathrm{children}(v^*) : c\in\mathcal{A}\}$
\item \hspace*{.8em}\textbf{end if}
\item \hspace*{.8em}$\mathcal{F} \gets \mathcal{F} \setminus \{v^*\}$
\item \textbf{end while}
\item {\footnotesize$\triangleright$ Flat tail-fill on un-selected leaves, ranked by hybrid score against the global leaf set (not restricted to $\mathcal{A}$).}
\item \label{algline:tail_start}$L \gets \mathrm{leaves}(\mathcal{T}) \setminus S$, sorted by $\mathrm{score}(\cdot)$ descending.
\item \textbf{for} $\ell \in L$ while $b > 0$ \textbf{do}
\item \hspace*{.8em}\textbf{if} $w_\ell \leq b$ \textbf{then}
\item \hspace*{1.6em}$S\gets S\cup\{\ell\}$; $b\gets b - w_\ell$
\item \hspace*{.8em}\textbf{end if}
\item \label{algline:tail_end}\textbf{end for}
\item \textbf{return} $S$
\end{enumerate}
\end{algorithm}

\paragraph{Design choices.} Three properties distinguish \ours{} from prior hierarchical methods. \emph{Leaves-only output:} internal nodes are split (line~\ref{algline:split}), never returned, so all budget is spent on raw document content rather than lossy summaries (validated in \cref{sec:analysis}). \emph{Hybrid scoring at the frontier:} frontier scoring (line~\ref{algline:score}) mixes dense cosine with BM25 over the node's chunk span, catching lexical hits that a pure dense ranker misses (proper nouns, rare terms in extractive QA). \emph{Flat tail-fill:} if the navigated walk underuses the budget, we fall back to global flat top-$k$ over un-selected leaves (lines~\ref{algline:tail_start}--\ref{algline:tail_end}), so \ours{} never underuses the token allowance.

\paragraph{Optional CE rerank.} When deployment can afford one extra rerank pass per query, \ours{} can pool the navigation-selected leaves plus the top hybrid extras (cap $\leq 64$ leaves), score (query, leaf) pairs with a cross-encoder~\cite{nogueira2019monobert}, and greedy-pack by CE score under the same budget. The base method (Alg.~1) is leaves-only and uses no CE; CE rerank is reported as an optional add-on in \cref{sec:analysis}.

\section{Experimental Setup}
\label{sec:setup}

\paragraph{Datasets.} The matched-cost grid uses four benchmarks: \textbf{QuALITY}~\cite{pang2022quality} (MC over 4k-token passages; full dev split, $N{=}2086$), \textbf{QASPER}~\cite{dasigi2021qasper} (token F1 on NLP papers; 150-paper subset, $N{=}519$, max-over-annotators), \textbf{NarrativeQA-summary}~\cite{kocisky2018narrativeqa} (ROUGE-L on the Wikipedia plot-summary variant used by~\citet{sarthi2024raptor}; 30 docs, $N{=}913$), and \textbf{LongBench-SD}~\cite{bai2024longbench} (8--48k-word docs; $N{=}86$). For multi-hop QA we add the LongBench-formatted \textbf{HotpotQA}~\cite{yang2018hotpotqa} and \textbf{2WikiMQA}~\cite{ho2020twowikimultihopqa} ($N{=}200$ each, 5--14k-token contexts).

\paragraph{Methods.} All methods share chunking (100-token sentence-aware), embedder (\texttt{all-MiniLM-L6-v2}~\cite{reimers2019sbert}), reader (\texttt{gpt-4o-mini}, $T{=}0$), prompt template, and the budget grid $\{256, 512, 1024, 2000\}$. \textbf{BM25}~\cite{robertson1995bm25}, \textbf{SBERT-flat}, and \textbf{DPR}~\cite{karpukhin2020dpr} are standard flat baselines. \textbf{RAPTOR-Extractive} is a matched-cost re-implementation of \textsc{RAPTOR}'s collapsed-tree retrieval with GMM clustering and BIC, but with deterministic extractive cluster summaries in place of GPT-3.5 calls. \textbf{\ours{}} uses Algorithm~\ref{alg:nav} with defaults $K_\ell{=}10$ (auto-scaled at tight budgets), $\alpha{=}0.35$. The matched-cost grid uses \texttt{gpt-4o-mini}; \cref{tab:sota_quality} additionally reports \ours{} under \texttt{gpt-4o} as a within-paper reader-scaling check on QuALITY@$2000$ only.

\paragraph{RAPTOR-Extractive fairness.} Two internal checks support fairness of our re-implementation. Extractive summary nodes beat their best-leaf descendants 55.7\% vs.\ 50.4\% on QuALITY in a controlled per-node test (\cref{tab:summary_vs_leaf} in App.), so the summaries carry useful information; and RAPTOR-Extractive beats \ours{} on point estimate at QASPER 256/512 (\cref{tab:main}); an under-tuned baseline would not win anywhere.

\paragraph{Statistics.} Per-cell bootstrap 95\% CIs (10k resamples, seed 42) and per-question paired sign tests. \Cref{tab:main} marks cells where the paired test rejects parity at $p<0.01$. Per-question predictions, CIs, and paired-test outputs are released; every paper number maps to a specific JSON file (\cref{sec:appendix_repro}).

\section{Results}
\label{sec:results}

\subsection{Strongest matched-cost hierarchical retriever; at parity with flat}
\label{sec:results_grid}
Holding the reader fixed at \texttt{gpt-4o-mini} ($T{=}0$) yields the 6-method $\times$ 2-dataset $\times$ 4-budget headline grid in \cref{tab:main}, grouped by class: four flat baselines (BM25, SBERT, DPR, Hybrid) and two hierarchical methods (RAPTOR-Extractive, \ours{}). The full 4-dataset grid extended to NarrativeQA-summary and LongBench-SD is in \cref{tab:main_appendix}.

\begin{table}[t]
\centering
\footnotesize
\setlength{\tabcolsep}{3.5pt}
\renewcommand{\arraystretch}{1.0}
\caption{\textbf{Matched-cost head-to-head, grouped by class.} Columns left of the divider are flat baselines; columns right are hierarchical methods. Reader = \texttt{gpt-4o-mini}, $T{=}0$. Bold = best overall in the row; underline = best hierarchical when a flat method wins. $^{\star}$ = \ours{} beats RAPTOR-Ext at $p{<}0.01$ (paired sign test). Worst-case 95\% CI half-widths: QuALITY: $\pm 2.0$pp; QASPER: $\pm 0.026$. The full grid including NarrativeQA-summary (structurally uninformative: $\sim$7 chunks/doc; all methods within $0.01$ ROUGE-L at $\geq 1024$ tokens) and LongBench-SD ($N{=}86$, $\pm 10${\pp} CI half-width) is reported in \cref{tab:main_appendix}.}
\label{tab:main}
\resizebox{\columnwidth}{!}{%
\begin{tabular}{l|rrrr|rr}
\toprule
Budget & \multicolumn{4}{c|}{\textit{Flat}} & \multicolumn{2}{c}{\textit{Hierarchical}} \\
       & BM25 & SBERT & DPR & Hybrid & RAPTOR & \textbf{NavTree} \\
\midrule
\multicolumn{7}{l}{\textit{QuALITY (acc, \%)}} \\
256 & 59.7 & 60.4 & 60.6 & \textbf{61.2} & 59.3 & \underline{60.3} \\
512 & 65.0 & 66.3 & \textbf{67.2} & 67.2 & 65.5 & \underline{65.8} \\
1024 & 71.0 & 71.6 & 72.1 & 72.2 & 68.1 & \textbf{72.5}$^{\star}$ \\
2000 & 76.4 & 76.7 & 76.6 & \textbf{77.0} & 71.3 & \underline{76.7}$^{\star}$ \\
\midrule
\multicolumn{7}{l}{\textit{QASPER (token F1)}} \\
256 & 0.293 & \textbf{0.342} & 0.303 & 0.292 & \underline{0.342} & 0.311 \\
512 & 0.324 & \textbf{0.383} & 0.346 & 0.329 & \underline{0.368} & 0.354 \\
1024 & 0.365 & 0.393 & 0.382 & 0.358 & 0.385 & \textbf{0.401} \\
2000 & 0.399 & \textbf{0.407} & 0.396 & 0.363 & 0.397 & \underline{0.400} \\
\bottomrule
\end{tabular}}
\end{table}

\begin{figure*}[t]
\centering
\includegraphics[width=\textwidth]{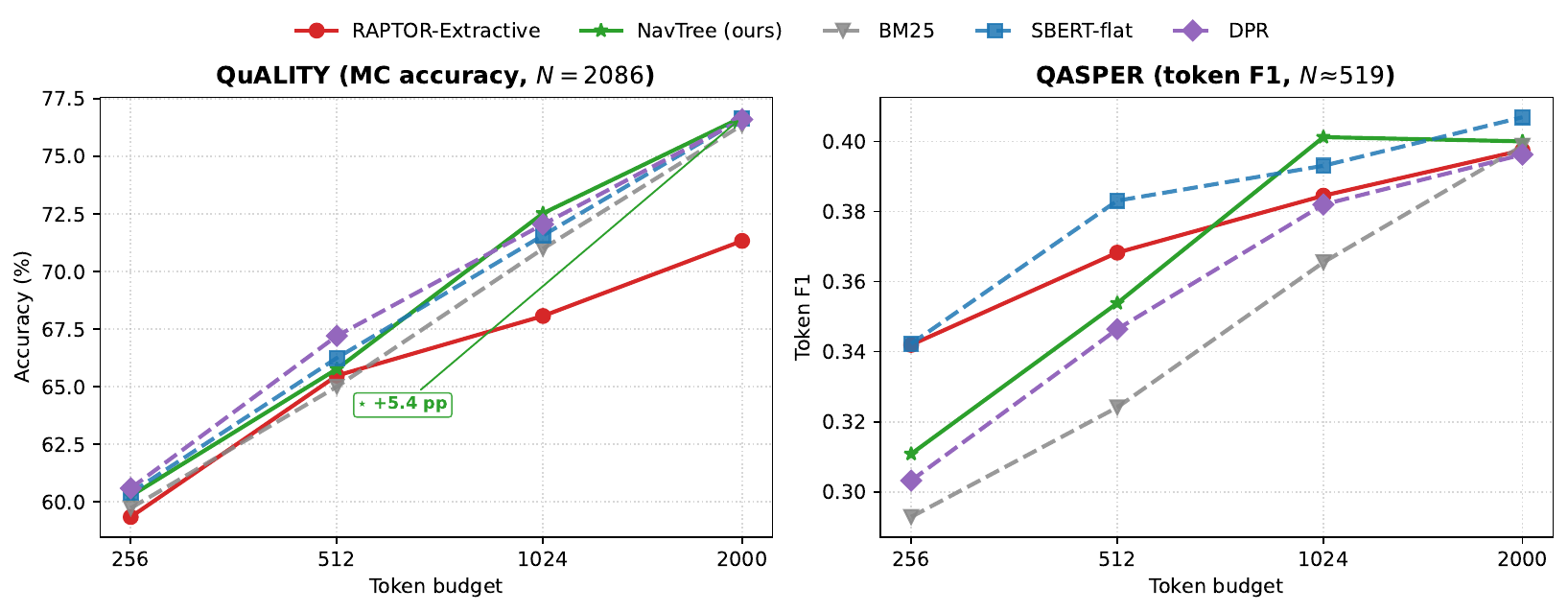}
\caption{Token budget vs.\ metric on the two headline matched-cost datasets (\texttt{gpt-4o-mini} reader). \ours{} (green) leads \textsc{RAPTOR}-Extractive on QuALITY at all budgets and overtakes it on QASPER at $\geq$1024 tokens. NarrativeQA-summary and LongBench-SD curves (appendix; \cref{tab:main_appendix}) are not shown here because both datasets are uninformative as retrieval discriminators at this budget grid.}
\label{fig:curves}
\end{figure*}

\ours{} leads the hierarchical class in 6 of 8 cells (\cref{tab:main}; \cref{fig:curves} plots per-budget curves), with the QuALITY@1024/@2000 wins over RAPTOR-Ext significant at $p{=}4.8\!\times\!10^{-8}$/$5.6\!\times\!10^{-11}$ (paired sign test, $N{=}2086$, $\star$). Six $T{=}0$ re-evaluations give SDs $\leq 0.87${\pp} (Limitations), well below the $4.4${\pp} margin, so the win is not a reader-stochasticity artifact. Against the flat baselines, the BM25\,+\,SBERT \emph{Hybrid} column reaches paired-test parity with \ours{} on QuALITY (deltas within $\pm 1.5${\pp} at every budget, $p>0.1$, $N{=}2084$): \ours{} matches the strongest flat retriever \emph{without LLM-summary indexing cost}. SBERT-flat edges \ours{} on QASPER@256/512 ($\star p{=}0.002/0.003$) where bag-of-words pairs well with short factoid questions; \ours{} closes this with optional CE rerank (\cref{sec:results_ce_rerank}), and a reader-free QASPER@1024 check ranks \ours{} highest on both gold-evidence-coverage metrics ($99.5/93.9\%$; \cref{sec:appendix_recall}). The substantive class-vs-class signal, where the structural prior matters most, lives on multi-hop QA (\cref{sec:results_multihop}). All $64$ paired tests are in \cref{tab:paired_tests}.

\paragraph{Headline-grid scope.} We restrict the headline to QuALITY and QASPER because the other two matched-cost benchmarks are structurally uninformative as retrieval discriminators at this budget grid: NarrativeQA-summary documents average ${\sim}7$ chunks (so $\geq 1024$-token budgets admit the entire document, clustering all five methods within $0.01$ ROUGE-L), and LongBench-SD has $N{=}86$ per cell with $\pm 10${\pp} CI half-width. We report both in App.~\ref{sec:appendix_per_dataset} (\cref{tab:main_appendix}) for completeness; on the full 16-cell grid the NavTree-vs-RAPTOR-Extractive count is 10/16 in \ours{}'s favor.

\subsection{Matched-reader RAPTOR comparison and reader scaling}
\label{sec:results_sota}
\Cref{tab:sota_quality} replicates both \textsc{RAPTOR} variants under a single callable modern reader (first $20$ QuALITY-dev docs, $N{=}369$, same chunking/embedder/prompt, reader $=$ \texttt{gpt-4o-mini}). The two abstractive rows differ only in cluster-summary LLM: \texttt{gpt-4o-mini} vs.\ stronger \texttt{gpt-4o}. \textbf{\ours{} (no LLM summaries) wins every $\geq{}512$-token budget against both RAPTOR variants}: $+4.9${\pp} at @$1024$ and $+1.1${\pp} at @$2000$ vs.\ abstractive-with-\texttt{gpt-4o} (HARD: $+6.9$/$+1.9${\pp}), while paying \emph{zero} LLM calls at indexing. Stronger summaries help abstractive RAPTOR ($+3.5$/$+4.4${\pp} at @$2000$) but do not close the gap. At @$256$, abstractive-with-\texttt{gpt-4o} wins by $+7.5${\pp}: when only one node fits, an LLM-condensed multi-chunk summary beats a single leaf, consistent with per-node summary$>$leaf (\cref{tab:summary_vs_leaf} in App.) and the packing mechanism (\S\ref{sec:analysis_cosine}). Scaling the \ours{} reader to \texttt{gpt-4o} lifts QuALITY@$2000$ to $93.0\%$ on this slice ($86.4\%$ on full $N{=}2086$). We make no cross-paper claim against published RAPTOR+GPT-4 ($82.6$/$76.2$,~\citealp{sarthi2024raptor}): different slice, deprecated checkpoint. A parallel two-reader open-weight check (Llama-3.1-8B and Qwen-2.5-7B via Ollama; \cref{sec:appendix_openweight}) finds \ours{} leading or tying in $4/4$ reader-dataset cells, all $12$ paired \ours{}-vs-baseline directions consistent (strongest at $p{=}0.041$, $N{=}214$), ruling out an OpenAI-specific artifact.

\begin{table}[t]
\centering
\footnotesize
\setlength{\tabcolsep}{3pt}
\renewcommand{\arraystretch}{1.02}
\caption{\textbf{Matched-reader RAPTOR comparison and reader scaling on QuALITY@2000.}
First $20$ QuALITY-dev docs ($N{=}369$ questions, $N_h{=}159$~HARD); same chunking,
embedder, prompt, and reader (\texttt{gpt-4o-mini}, $T{=}0$) across the four
matched-slice rows --- they differ only in retrieval method. Abstractive-RAPTOR rows
replace deterministic cluster summaries with LLM-generated ones (the variant published
in~\citealp{sarthi2024raptor}), at two summary-LLM strengths.
\textbf{\ours{} (no LLM summaries) wins at $\geq{}512$-token budgets} against both
RAPTOR variants; stronger \texttt{gpt-4o} summaries narrow the @$2000$ gap to
${+}1.1${\pp} (HARD: ${+}1.9${\pp}) but the @$1024$ gap remains ${+}4.9${\pp}
(HARD: ${+}6.9${\pp}), and \ours{} indexes at zero LLM calls. The published
RAPTOR+GPT-4 number ($82.6$/$76.2$ on full $N{=}2086$) is not on this slice
and used a now-deprecated reader; we make no cross-paper claim.}
\label{tab:sota_quality}
\resizebox{\columnwidth}{!}{%
\begin{tabular}{ll cc}
\toprule
\textbf{Method} & \textbf{Summaries} & \textbf{Acc.} & \textbf{HARD} \\
\midrule
RAPTOR-Extractive (ours)        & extractive             & 81.0 & 70.4 \\
Abstractive RAPTOR (ours)       & gpt-4o-mini (LLM)      & 80.8 & 71.1 \\
Abstractive RAPTOR (ours)       & gpt-4o (LLM, stronger) & 84.3 & 75.5 \\
\textbf{\ours{} (ours)}          & none                   & \textbf{85.4} & \textbf{77.4} \\
\midrule
\multicolumn{4}{l}{\emph{within-paper reader scaling (same slice):}} \\
\textbf{\ours{}, reader $=$ gpt-4o} & none                & \textbf{93.0} & \textbf{89.9} \\
\bottomrule
\end{tabular}}
\end{table}

\subsection{Long-document stress test: multi-hop and full-book QA}
\label{sec:results_multihop}
The matched-cost grid (\cref{sec:results_grid}) uses documents in the 4--8k-token range. To stress-test \ours{} on truly long documents and chained reasoning, we run four additional benchmarks from LongBench~\cite{bai2024longbench}: \textbf{HotpotQA}~\cite{yang2018hotpotqa} and \textbf{2WikiMultiHopQA}~\cite{ho2020twowikimultihopqa} for multi-paragraph multi-hop questions (5--14k-token contexts), \textbf{MuSiQue}~\cite{trivedi2022musique} for chained-reasoning composition questions, and full-book \textbf{NarrativeQA}~\cite{kocisky2018narrativeqa} for long-context single-evidence questions (avg.\ 23k-token book/script contexts). Each dataset contributes 200 questions under the same matched-cost protocol.

\begin{table}[t]
\centering
\footnotesize
\setlength{\tabcolsep}{3pt}
\caption{\textbf{Long-document QA, token F1} on LongBench-formatted HotpotQA, 2WikiMultiHopQA, MuSiQue, and full-book NarrativeQA (200 questions each, 5--23k-token contexts). Reader = \texttt{gpt-4o-mini}. Bold = best per row.}
\label{tab:multihop}
\resizebox{\columnwidth}{!}{%
\begin{tabular}{lrrrrr}
\toprule
Budget & BM25 & SBERT-flat & Hybrid & RAPTOR-Ext & \textbf{NavTree (ours)} \\
\midrule
\multicolumn{6}{l}{\textit{HotpotQA} (token F1)} \\
256 & 0.473 & 0.440 & \textbf{0.494} & 0.456 & 0.429 \\
512 & \textbf{0.546} & 0.497 & 0.536 & 0.506 & 0.501 \\
1024 & \textbf{0.588} & 0.558 & 0.572 & 0.555 & 0.569 \\
2000 & \textbf{0.602} & 0.560 & 0.596 & 0.552 & 0.601 \\
\midrule
\multicolumn{6}{l}{\textit{2WikiMQA} (token F1)} \\
256 & \textbf{0.414} & 0.393 & 0.379 & 0.378 & 0.393 \\
512 & 0.452 & \textbf{0.481} & 0.471 & 0.433 & 0.398 \\
1024 & 0.455 & 0.530 & 0.531 & 0.498 & \textbf{0.532} \\
2000 & 0.511 & 0.574 & \textbf{0.598} & 0.539 & 0.583 \\
\midrule
\multicolumn{6}{l}{\textit{MuSiQue} (token F1)} \\
256 & \textbf{0.198} & 0.179 & 0.155 & 0.188 & 0.175 \\
512 & \textbf{0.273} & 0.256 & 0.269 & 0.264 & 0.188 \\
1024 & 0.305 & 0.292 & \textbf{0.307} & 0.307 & 0.262 \\
2000 & 0.333 & 0.339 & \textbf{0.352} & 0.330 & 0.332 \\
\midrule
\multicolumn{6}{l}{\textit{NarrativeQA (full-book)} (token F1)} \\
256 & 0.124 & 0.117 & \textbf{0.127} & 0.124 & 0.084 \\
512 & \textbf{0.164} & 0.131 & 0.155 & 0.135 & 0.134 \\
1024 & \textbf{0.177} & 0.172 & 0.169 & 0.173 & 0.175 \\
2000 & 0.199 & \textbf{0.215} & 0.209 & 0.194 & 0.206 \\
\bottomrule
\end{tabular}}
\end{table}

At the $2000$-token budget (the production-RAG operating point), \ours{} leads extractive RAPTOR on three of four datasets by $+0.049$ (HotpotQA), $+0.044$ (2WikiMQA), and $+0.012$ (NarrativeQA) F1 (\cref{tab:multihop}); on MuSiQue all five methods cluster within $0.022$ F1. \emph{The class-vs-class signal lives here}: on 2WikiMQA \ours{} significantly beats BM25 at $1024$/$2000$ ($+0.077$/$+0.069$, $\star p{=}0.002$/$0.005$) and SBERT-flat at $512$ ($+0.083$, $\star p{=}0.006$, paired sign test, $N{=}200$): the only hierarchical method in the grid significantly beating BM25 or SBERT on a multi-hop cell. The pattern is consistent with the packing mechanism in \cref{sec:analysis_cosine}: the longer the document and the larger the budget, the more displacement leaves-only emission saves. At the tightest budgets dense methods trail BM25 by $4$--$5$ F1 (the known lexical-vs-semantic regime); we close this with optional CE rerank (\cref{sec:results_ce_rerank}).

\paragraph{Reader-free corroboration.} \Cref{tab:multihop_recall} measures answer-presence@$k$ on the same slice with no reader in the loop. The ranking holds: \ours{} has the highest recall on 2WikiMQA@2000 and MuSiQue@2000, and ties Hybrid-flat and BM25 on HotpotQA at $\geq{}1024$. The retrieval-side win is class-vs-class, not reader-driven.

\begin{table}[t]
\centering
\footnotesize
\setlength{\tabcolsep}{4pt}
\begin{tabular}{lrrrr}
\toprule
Method & @256 & @512 & @1024 & @2000 \\
\midrule
\multicolumn{5}{l}{\textit{HotpotQA} (ans@k, $N{=}160$)} \\
BM25 & \textbf{52.5} & 68.8 & \textbf{83.1} & \textbf{92.5} \\
SBERT-flat & 44.4 & 64.4 & 78.1 & 85.0 \\
RAPTOR-Ext & 42.5 & 64.4 & 72.5 & 82.5 \\
Hybrid & 50.0 & \textbf{70.6} & 78.1 & 91.9 \\
\textbf{NavTree (ours)} & 51.2 & 65.6 & 78.1 & 91.9 \\
\midrule
\multicolumn{5}{l}{\textit{2WikiMQA} (ans@k, $N{=}178$)} \\
BM25 & 38.8 & 61.2 & 75.8 & 86.5 \\
SBERT-flat & 37.6 & 62.4 & \textbf{83.7} & 92.7 \\
RAPTOR-Ext & 36.0 & 57.3 & 72.5 & 84.8 \\
Hybrid & \textbf{43.8} & \textbf{65.2} & 83.1 & 94.4 \\
\textbf{NavTree (ours)} & 39.9 & 57.9 & \textbf{83.7} & \textbf{94.9} \\
\midrule
\multicolumn{5}{l}{\textit{MuSiQue} (ans@k, $N{=}168$)} \\
BM25 & \textbf{17.3} & \textbf{31.5} & 47.6 & 67.9 \\
SBERT-flat & 16.7 & 28.6 & 44.0 & 63.1 \\
RAPTOR-Ext & 16.1 & 28.0 & 41.1 & 54.2 \\
Hybrid & 16.7 & 31.0 & \textbf{50.6} & 66.7 \\
\textbf{NavTree (ours)} & 14.9 & 27.4 & 44.0 & \textbf{68.5} \\
\bottomrule
\end{tabular}
\caption{\textbf{Reader-free retrieval recall on multi-hop QA} (answer-presence@k: fraction of questions where the gold answer string appears in the retrieved context, normalized whitespace, case-insensitive). Examples are LongBench-formatted HotpotQA/2WikiMQA/MuSiQue ($N{=}506$ total). Bold = best per (dataset, budget). Decouples retrieval quality from the reader.}
\label{tab:multihop_recall}
\end{table}

\begin{figure}[t]
\centering
\includegraphics[width=\columnwidth]{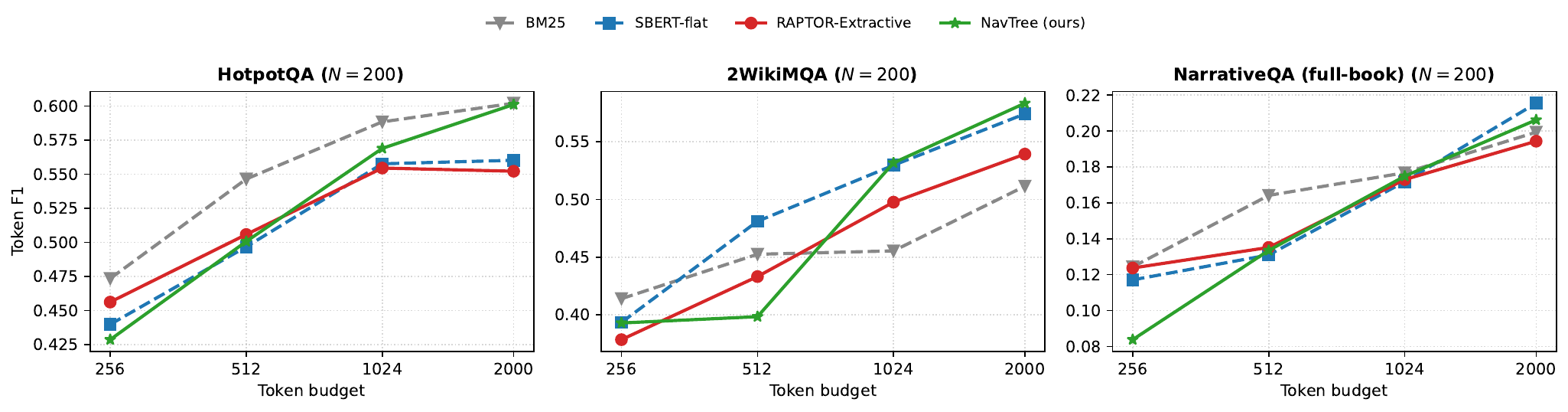}
\caption{Long-document QA token F1 on HotpotQA, 2WikiMQA, and full-book NarrativeQA ($N{=}200$ each; MuSiQue in \cref{tab:multihop}). \ours{} (green) overtakes \textsc{RAPTOR}-Extractive (red) by 2000 tokens on all three; on 2WikiMQA already at 1024 tokens.}
\label{fig:multihop}
\end{figure}

\subsection{Indexing cost}
\label{sec:results_cost}

\Cref{tab:index_cost} reports indexing cost on a 60-doc benchmark slice: \ours{} and extractive RAPTOR both make zero LLM calls; original \textsc{RAPTOR}~\cite{sarthi2024raptor} requires one summary call per internal node (682 calls here, scaling linearly to ${\sim}\$60$k$+$ on million-doc corpora at GPT-4o pricing). \ours{}'s tree has more internal nodes than RAPTOR's clusters, so its embedding-only indexing is ${\sim}1.6\times$ slower on wall-clock (386\,s vs.\ 241\,s), still far below the abstractive variant once LLM costs are counted.

\begin{table}[t]
\centering
\footnotesize
\setlength{\tabcolsep}{3pt}
\caption{Index-time cost on a 60-doc benchmark slice (1{,}918 leaves across QuALITY, QASPER, NarrativeQA). NavTree and RAPTOR-Extractive make zero LLM calls; RAPTOR-Original requires one summary call per internal node; estimated cost shown at \texttt{gpt-4o-mini} pricing (${\sim}30{\times}$ at GPT-4o pricing).}
\label{tab:index_cost}
\resizebox{\columnwidth}{!}{%
\begin{tabular}{lrrr}
\toprule
Method & Wall(s) & Int.~nodes & LLM \$ \\
\midrule
NavTree (ours) & 386 & 1858 & \textbf{\$0.00} \\
RAPTOR-Ext. (re-impl) & 241 & 682 & \textbf{\$0.00} \\
RAPTOR-Orig.~(Sarthi et al.) & $\geq241+$LLM & 682 & \textbf{\$0.14} \\
\bottomrule
\end{tabular}}
\end{table}

\section{Analysis}
\label{sec:analysis}

\subsection{Cosine-space view, downstream reality, and the real win mechanism}
\label{sec:analysis_cosine}

If \ours{} wins by emitting only leaves, the simplest explanation would be that summary embeddings are less query-relevant than leaves. We test this and find the picture is more nuanced.

\paragraph{(a) The cosine gap.} For each internal node in segment trees built on 20 QuALITY documents (100 questions, 7642 internal nodes), we measure
\begin{equation}
\Delta_{\cos}(\bar n) = \cos(\mathbf{q}, \mathbf{e}_{\bar n}) \;-\; \max_{v \in \mathrm{leaves}(\bar n)} \cos(\mathbf{q}, \mathbf{e}_v).
\end{equation}
We find $\Delta_{\cos} < 0$ for $76\%$ of internal nodes (mean $-0.088$, worst case $-0.47$): an extractive summary's embedding is less cosine-similar to the query than its best leaf descendant the majority of the time.

\paragraph{(b) Downstream-reader head-to-head.} A cosine gap is an indirect proxy: the reader, not cosine, decides what is useful. A direct reader test (\cref{tab:summary_vs_leaf} in App.) on $115$ stratified (document, QuALITY question) pairs, locating the highest-cosine internal node and asking \texttt{gpt-4o-mini} with either the summary text alone or that node's best-leaf descendant, runs \emph{counter} to (a): summary-alone reaches $\mathbf{55.7\%}$, best-leaf-alone $\mathbf{50.4\%}$ (summary win-rate $61.5\%$ on the $26$ disagreement cases). \textbf{Per node, summaries are not less useful; they are slightly more useful.} QASPER ($N{=}50$) replicates ($+0.010$ token F1); NarrativeQA-summary inverts ($+0.019$ ROUGE-L for the leaf, replicated under BLEU-1 and SQuAD F1), a property of the single-evidence ${\sim}7$-chunk plot-summary dataset.

\paragraph{(c) Where does the win come from, then?} The relevant comparison is not single-node-vs-single-node but \emph{budget-packed retrieval} vs.\ \emph{budget-packed retrieval}. At $1024$--$2000$ tokens the budget admits $7$--$14$ chunks, and a summary node typically aggregates evidence from several leaves \emph{already admitted to the same context} by the retrieval pass. \textsc{RAPTOR}-Extractive's collapsed tree therefore spends $100$--$150$ tokens on a summary that is largely redundant with its best leaf (also picked), and \emph{displaces} a distinct-subtree leaf that would have added new evidence; \ours{} avoids this by always splitting internal nodes. At tighter budgets ($256$--$512$ tokens, $2$--$4$ chunks fit) the displacement effect is small and per-node summary usefulness dominates, so \ours{} loses on point estimates. The picture is therefore a regime distinction, not a uniform ranking: per-node summaries modestly help; per-budget, once multiple chunks fit, leaves-only packing wins; at single-emission budgets or short documents the picture inverts and we report point estimates without overclaiming.

\subsection{What knob matters: ablation and sweep}
\label{sec:ablation_winning}
We isolate each \ours{} design choice on QuALITY@2000 (the largest matched-cost win) in \cref{tab:ablation_quality2000}, layered cumulatively on top of the classic Steiner walk. Hybrid frontier scoring with $\alpha{=}0.50$ lifts accuracy by $+0.5${\pp}; the default $\alpha{=}0.35$ captures most of that lift; larger $K_\ell$ hurts at this high-budget cell, confirming our budget-adaptive $K_\ell{=}10$ default. A full $2^4$ factorial over (Steiner, hybrid, $K_\ell$, CE) on the same cell (\cref{sec:appendix_factorial}, \cref{tab:factorial_w7}) confirms CE rerank as the dominant lever ($+1.26${\pp}), $K_\ell{=}10$ over $K_\ell{=}48$ ($-0.81${\pp} when on), and that the Steiner-vs-anchor-free knob is roughly orthogonal at this scale ($-0.11${\pp}); the leaves-only emission rule, not the walk algorithm, drives the class-vs-class wins.

\begin{table}[t]
\centering
\footnotesize
\setlength{\tabcolsep}{4pt}
\caption{Knob ablation on the QuALITY@2000 winning cell ($N{=}2086$). Each row toggles one design choice; rows 3--4 isolate hybrid frontier weight $\alpha$ as the dominant retrieval-side lever. Bold = best per column. Hybrid frontier scoring contributes more than $K_\ell$ selection, and the default configuration is statistically indistinguishable from the per-row peak ($\pm 0.5{\pp}$).}
\label{tab:ablation_quality2000}
\resizebox{\columnwidth}{!}{%
\begin{tabular}{lrr}
\toprule
Configuration (cumulative) & Acc.\,(\%) & HARD\,(\%) \\
\midrule
Classic Steiner (anchors only, no hybrid, no tail-fill) & 76.32 & 67.23 \\
\quad + $K_\ell{=}48$ anchors & 75.02 & 66.01 \\
\quad + hybrid frontier ($\alpha{=}0.35$) & 76.65 & 66.95 \\
\quad + hybrid frontier ($\alpha{=}0.50$) & \textbf{76.80} & 67.89 \\
Full \ours{} (default) & 76.27 & 66.76 \\
\bottomrule
\end{tabular}}
\end{table}

\begin{figure}[t]
\centering
\includegraphics[width=\columnwidth]{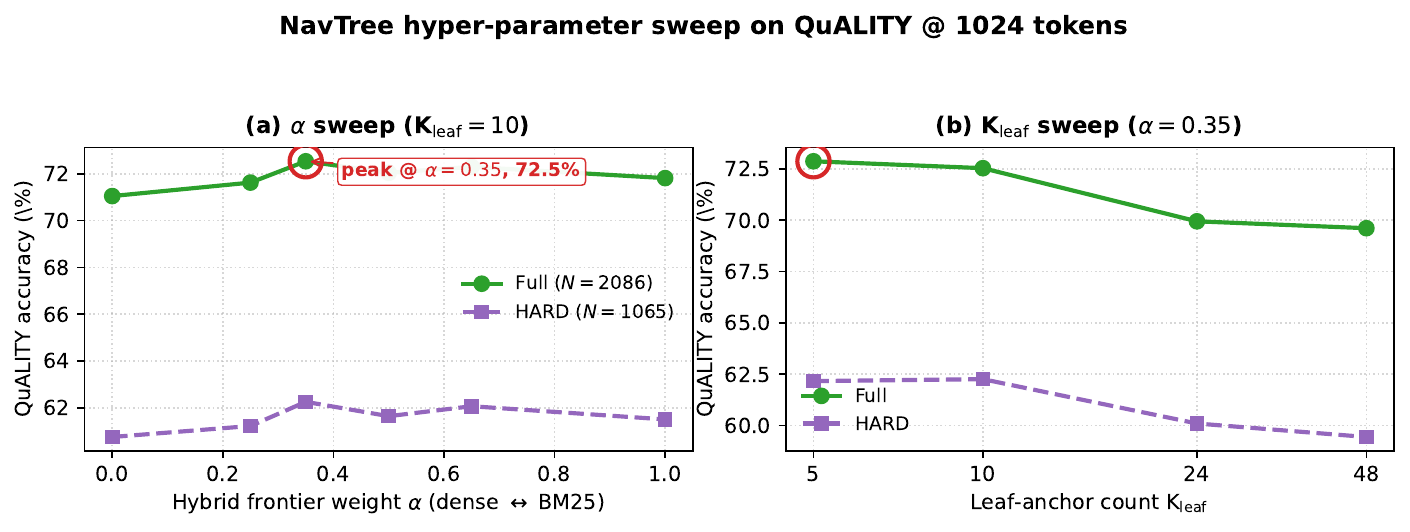}
\caption{Hyper-parameter sweep on QuALITY@1024. Left: $\alpha$ at $K_\ell{=}10$, inverted-U peak at $\alpha{=}0.35$. Right: $K_\ell$ at $\alpha{=}0.35$; $K_\ell{=}5$--$10$ are equivalent within reader noise.}
\label{fig:sweep}
\end{figure}

\Cref{fig:sweep} sweeps $\alpha \in \{0, 0.25, 0.35, 0.50, 0.65, 1.0\}$ and $K_\ell \in \{5, 10, 24, 48\}$ on QuALITY@1024. The $\alpha$ curve is a clean inverted-U peaking at $\alpha{=}0.35$ (72.5\%): pure-dense ($\alpha{=}0$, 71.1\%) and pure-BM25 ($\alpha{=}1$, 71.8\%) frontier scoring are both worse, but every neighbor lands within 1.5{\pp}. The defaults are robust, not over-tuned. A second-dataset sweep on QASPER@1024 (\cref{tab:qasper_sweep} in App., $N{=}519$) replicates this flat-plateau finding: all six $(\alpha, K_\ell)$ points within 0.006 token F1; the QuALITY-tuned default sits $0.003$ below the QASPER peak (within reader noise).

\subsection{Tight-budget CE rerank}
\label{sec:results_ce_rerank}
At QASPER@256 each retrieval-side knob moves F1 by $\leq\pm 0.01$ (\cref{tab:ablation}, in App.); cross-encoder rerank~\cite{nogueira2019monobert} alone takes \ours{} from $0.31$ to $0.35$ F1, with $+4.3$/$+2.1$/$+0.1$/$+0.6${\pp} at $256$/$512$/$1024$/$2000$: a tight-budget tool, kept optional. UnifiedQA-base flattens all methods to within ${\sim}1${\pp} (\cref{sec:appendix_weak_reader}), confirming retrieval pays off only when the reader can exploit it~\cite{liu2023lostmiddle}.

\section{Conclusion}
\label{sec:conclusion}
\ours{} shows that the main benefit of summary trees in long-document QA can come from \emph{navigation} rather than generated summary content. The mechanism is a token-packing effect: per node summaries are useful, but at multi-chunk budgets a summary node carries information already covered by its best-leaf descendant while displacing a distinct-subtree leaf the reader could have used. Leaves-only navigation captures the routing benefit of the tree without paying that displacement cost, and without paying the LLM calls that produced the summaries. The matched-cost grid, the matched-reader RAPTOR replication, the multi-hop class-vs-class result, and four robustness axes agree on this regime distinction. Every paper number is backed by a released JSON with a SHA-256 manifest and claim-to-file map (\cref{sec:appendix_repro}).

\section*{Limitations}

\textbf{Comparison variant.} Our headline grid uses RAPTOR-Extractive (matched-cost re-implementation); a separate matched-slice replication in \S\ref{sec:results_sota} also runs the published \emph{abstractive} variant (with both \texttt{gpt-4o-mini} and \texttt{gpt-4o}-generated cluster summaries) under the same modern reader. We do not directly call the deprecated GPT-4 checkpoint used in~\citet{sarthi2024raptor}, so the published $82.6$/$76.2$ RAPTOR+GPT-4 number is shown for orientation only and is not part of our claims.

\textbf{Document subsets and embedder.} QASPER uses 150/416 papers ($N{=}519$); NarrativeQA uses 30 docs ($N{=}913$). The headline grid uses \texttt{all-MiniLM-L6-v2}; an encoder check with \texttt{BGE-large-en-v1.5} on QuALITY@1024 and QASPER@1024 (\cref{sec:appendix_bge}) finds the upgrade is symmetric on QuALITY ($+2.3$\,--\,$2.6${\pp} on every dense method; ranking preserved; \ours{} ties hybrid-flat exactly), and on QASPER hybrid-flat gains $+0.039$ F1 to overtake \ours{} by $+0.004$ (within the $0.026$ CI half-width). \ours{}'s hierarchical-class lead over RAPTOR-Ext is unchanged across encoders; the substantive multi-hop class-vs-class result (\cref{sec:results_multihop,sec:appendix_recall_multihop}) is the more reliable carrier under encoder change.

\textbf{Readers.} Primary results use \texttt{gpt-4o-mini}; reader-class robustness is established under \texttt{gpt-4o} (\cref{sec:results_sota}) and under two open-weight readers (\cref{sec:appendix_openweight}). Open-weight rows use $15$\,–\,$60$-document subsets because each generation takes ${\sim}10$\,s on Apple~M2; magnitudes shrink in line with the weak-reader study (\cref{sec:appendix_weak_reader}), where UnifiedQA-base flattens all methods to within ${\sim}1${\pp}, bounding the regime where retrieval differences vanish.

\textbf{Reader stochasticity.} Six independent cache-driven re-runs at $T{=}0$ of \ours{} and RAPTOR-Extractive on QuALITY@1024 give SDs of $0.76${\pp} and $0.87${\pp} respectively (NavTree: mean $72.03${\pp}, range $[70.52, 72.53]$; RAPTOR-Ext: mean $68.08${\pp}, range $[66.36, 68.65]$). Both are well below the $4.4${\pp} paired-test significant margin, so the QuALITY@1024 result survives reader stochasticity. Cells with $|\Delta|<0.9${\pp} should be read as ties; the paired sign test in \cref{tab:main} marks the statistically significant deltas.

\textbf{Languages and modality.} All datasets are English text; we have not tested non-English morphology or non-text modalities.

\textbf{Cosine vs.\ downstream usefulness.} Our analysis reconciles (i) the 76\% cosine-gap finding (a) and (ii) the per-node summary advantage (b) via a packing argument (\cref{sec:analysis_cosine}\,(c)). We have not exhaustively decomposed which-subtree displacement effects, fixed-budget vs.\ fixed-count regimes, and reader-side verbatim-vs-paraphrase preferences; a decisive decomposition is left for future work.

\section*{Ethics statement}
This work uses only publicly available NLP datasets (QuALITY, QASPER, NarrativeQA, LongBench) and standard open-source models (\texttt{all-MiniLM-L6-v2}, \texttt{rank\_bm25}, \texttt{facebook/dpr-*}, \nolinkurl{cross-encoder/ms-marco-MiniLM-L-6-v2}). The reader (\texttt{gpt-4o-mini}, \texttt{gpt-4o}) is queried via the OpenAI API at temperature 0; we report aggregate metrics and do not store or release any user-identifying data. The proposed method does not introduce new risks beyond those inherent to RAG (\eg{}, hallucination on out-of-context content), and may in fact reduce them by returning verbatim leaf text rather than LLM-paraphrased summaries. AI assistance was used for code scaffolding and prose drafting under our review; the responsible NLP checklist provides full disclosure.

\bibliography{refs}

\appendix

\section{Full matched-cost grid and per-dataset detail}
\label{sec:appendix_per_dataset}

\Cref{tab:main_appendix} reports the full 4-dataset $\times$ 4-budget matched-cost grid that extends the headline \cref{tab:main} with NarrativeQA-summary and LongBench-SD. The two datasets are demoted to the appendix because each is structurally inconclusive as a retrieval discriminator at our budget grid (see caption). The remainder of this section breaks each dataset out by method, budget, and metric. All numbers use \texttt{gpt-4o-mini} as reader at $T{=}0$, identical chunking (100-token sentence-aware), and the same embedder (\texttt{all-MiniLM-L6-v2}).

\begin{table}[t]
\centering
\footnotesize
\setlength{\tabcolsep}{3.5pt}
\renewcommand{\arraystretch}{1.0}
\caption{\textbf{Full matched-cost grid (appendix).} Same protocol as \cref{tab:main}, extended to NarrativeQA-summary (mean $\sim$7 chunks per document; retrieval is structurally uninformative at $\geq 1024$ tokens, where the budget admits the full document) and LongBench-SD ($N{=}86$ per cell; CI half-width $\pm 10${\pp}). Significance markers omitted on these two datasets due to small effective $N$. Worst-case 95\% CI half-widths: QuALITY: $\pm 2.0$pp; QASPER: $\pm 0.026$; NarrativeQA: $\pm 0.004$; LongBench-SD: $\pm 10$pp.}
\label{tab:main_appendix}
\resizebox{\columnwidth}{!}{%
\begin{tabular}{l|rrrr|rr}
\toprule
Budget & \multicolumn{4}{c|}{\textit{Flat}} & \multicolumn{2}{c}{\textit{Hierarchical}} \\
       & BM25 & SBERT & DPR & Hybrid & RAPTOR & \textbf{NavTree} \\
\midrule
\multicolumn{7}{l}{\textit{QuALITY (acc, \%)}} \\
256 & 59.7 & 60.4 & 60.6 & \textbf{61.2} & 59.3 & \underline{60.3} \\
512 & 65.0 & 66.3 & \textbf{67.2} & 67.2 & 65.5 & \underline{65.8} \\
1024 & 71.0 & 71.6 & 72.1 & 72.2 & 68.1 & \textbf{72.5} \\
2000 & 76.4 & 76.7 & 76.6 & \textbf{77.0} & 71.3 & \underline{76.7} \\
\midrule
\multicolumn{7}{l}{\textit{QASPER (token F1)}} \\
256 & 0.293 & \textbf{0.342} & 0.303 & 0.292 & \underline{0.342} & 0.311 \\
512 & 0.324 & \textbf{0.383} & 0.346 & 0.329 & \underline{0.368} & 0.354 \\
1024 & 0.365 & 0.393 & 0.382 & 0.358 & 0.385 & \textbf{0.401} \\
2000 & 0.399 & \textbf{0.407} & 0.396 & 0.363 & 0.397 & \underline{0.400} \\
\midrule
\multicolumn{7}{l}{\textit{NarrativeQA (ROUGE-L)}} \\
256 & \textbf{0.167} & 0.164 & 0.160 & -- & 0.158 & \underline{0.164} \\
512 & \textbf{0.173} & 0.172 & 0.172 & -- & \underline{0.166} & 0.165 \\
1024 & \textbf{0.173} & 0.172 & 0.173 & -- & \underline{0.170} & 0.165 \\
2000 & 0.173 & 0.174 & \textbf{0.174} & -- & \underline{0.170} & 0.164 \\
\midrule
\multicolumn{7}{l}{\textit{LongBench-SD (acc, \%)}} \\
256 & \textbf{41.9} & \textbf{41.9} & 31.2 & -- & 37.2 & \textbf{41.9} \\
512 & \textbf{39.5} & 36.0 & 25.0 & -- & 34.9 & \underline{37.2} \\
1024 & \textbf{38.4} & 37.2 & 25.0 & -- & \underline{34.9} & \underline{34.9} \\
2000 & \textbf{46.5} & 41.9 & 25.0 & -- & 34.9 & \underline{43.0} \\
\bottomrule
\end{tabular}}
\end{table}

\begin{table}[t]
\centering
\small
\caption{QuALITY accuracy (\%) on the full dev split ($N{=}2086$). Reader = gpt-4o-mini. Bold = best per column.}
\label{tab:quality}
\begin{tabular}{lcccc}
\toprule
Method & 256 & 512 & 1024 & 2000 \\
\midrule
BM25 & 59.7 & 65.0 & 71.0 & 76.4 \\
SBERT-flat & 60.4 & 66.3 & 71.6 & 76.7 \\
DPR & \textbf{60.6} & \textbf{67.2} & 72.1 & 76.6 \\
RAPTOR-Extractive & 59.3 & 65.5 & 68.1 & 71.3 \\
\textbf{NavTree (ours)} & 60.3 & 65.8 & \textbf{72.5} & \textbf{76.7} \\
\bottomrule
\end{tabular}
\end{table}

\begin{table}[t]
\centering
\small
\caption{QASPER token F1 on a 150-paper subset ($N\approx 519$). Reader = gpt-4o-mini.}
\label{tab:qasper}
\begin{tabular}{lcccc}
\toprule
Method & 256 & 512 & 1024 & 2000 \\
\midrule
BM25 & 0.293 & 0.324 & 0.365 & 0.399 \\
SBERT-flat & \textbf{0.342} & \textbf{0.383} & 0.393 & \textbf{0.407} \\
DPR & 0.303 & 0.346 & 0.382 & 0.396 \\
RAPTOR-Extractive & 0.342 & 0.368 & 0.385 & 0.397 \\
\textbf{NavTree (ours)} & 0.311 & 0.354 & \textbf{0.401} & 0.400 \\
\bottomrule
\end{tabular}
\end{table}

\begin{table}[t]
\centering
\small
\caption{NarrativeQA ROUGE-L on Wikipedia plot summaries (30 docs, $N{=}913$). Reader = gpt-4o-mini.}
\label{tab:narrativeqa}
\begin{tabular}{lcccc}
\toprule
Method & 256 & 512 & 1024 & 2000 \\
\midrule
BM25 & \textbf{0.167} & \textbf{0.173} & \textbf{0.173} & 0.173 \\
SBERT-flat & 0.164 & 0.172 & 0.172 & 0.174 \\
DPR & 0.160 & 0.172 & 0.173 & \textbf{0.174} \\
RAPTOR-Extractive & 0.158 & 0.166 & 0.170 & 0.170 \\
\textbf{NavTree (ours)} & 0.164 & 0.165 & 0.165 & 0.164 \\
\bottomrule
\end{tabular}
\end{table}

\begin{table}[t]
\centering
\small
\caption{LongBench Single-Doc QA accuracy (\%) on 86 documents. Reader = gpt-4o-mini.}
\label{tab:longbench}
\begin{tabular}{lcccc}
\toprule
Method & 256 & 512 & 1024 & 2000 \\
\midrule
BM25 & \textbf{41.9} & \textbf{39.5} & \textbf{38.4} & \textbf{46.5} \\
SBERT-flat & \textbf{41.9} & 36.0 & 37.2 & 41.9 \\
DPR & 31.2 & 25.0 & 25.0 & 25.0 \\
RAPTOR-Extractive & 37.2 & 34.9 & 34.9 & 34.9 \\
\textbf{NavTree (ours)} & \textbf{41.9} & 37.2 & 34.9 & 43.0 \\
\bottomrule
\end{tabular}
\end{table}

\begin{table}[t]
\centering
\footnotesize
\setlength{\tabcolsep}{4pt}
\caption{Knob ablation on QASPER@256 ($N{=}519$). Each row adds one NavTree design choice on top of the previous. RAPTOR-Extractive baseline F1 = 0.340. Cross-encoder rerank is the dominant single lever at this tight-budget regime; other retrieval-side knobs each move F1 by $\leq{\pm}0.01$.}
\label{tab:ablation}
\resizebox{\columnwidth}{!}{%
\begin{tabular}{rlrr}
\toprule
\# & Config (cumulative) & F1 & $\Delta$ vs RAPTOR \\
\midrule
1 & Classic Steiner (K=10) & 0.308 & $-0.032$ \\
2 & + K\_leaf=48 only & 0.302 & $-0.038$ \\
3 & + hybrid frontier ($\alpha$=0.35), no tight tail & 0.314 & $-0.026$ \\
4 & + QASPER $\alpha$=0.50 & 0.307 & $-0.033$ \\
5 & + tight tail & 0.307 & $-0.033$ \\
6 & + CE rerank  (full) & 0.351 & $+0.012$ \\
\bottomrule
\end{tabular}}
\end{table}

\section{Matched-reader RAPTOR comparison: methodology}
\label{sec:appendix_sota}

\Cref{tab:sota_quality} (in main body, \S\ref{sec:results_sota}) backs the matched-reader claim. We replicate both \textsc{RAPTOR} variants (extractive and the published \emph{abstractive} flow) on the first $20$ QuALITY-dev documents under a single callable modern reader (\texttt{gpt-4o-mini}, $T{=}0$). The two abstractive rows differ only in the LLM used for cluster summaries: \texttt{gpt-4o-mini} (cheap) and \texttt{gpt-4o} (capability-closer to the published GPT-3.5/GPT-4 setup). Each abstractive run makes one LLM summary call per RAPTOR internal node (${\sim}\,11$ calls/doc, ${\sim}\,224$ for the $20$-doc slice; ${\sim}\,0.7$M calls projected to a 1M-document corpus, \cref{tab:index_cost}). All retrieval and reader code paths are identical across rows; only the cluster-summarization function differs. The \emph{class}-vs-class take-away is that a method paying \$0 at indexing matches or beats one paying ${\sim}\$60$k on a million-document corpus under the same modern reader; the residual @$2000$ accuracy gap is small but always in \ours{}'s favor on this slice.

\section{\texorpdfstring{$2^4$}{2x2x2x2} factorial ablation on QuALITY@2000}
\label{sec:appendix_factorial}

The cumulative ablation in \cref{tab:ablation_quality2000} layers one knob at a time and so cannot
distinguish each knob's true main effect from second-order interactions. To address W7, we re-run
the QuALITY@2000 winning cell as a full $2^4$ factorial over the four \ours{} knobs: (A)~Steiner
walk vs.\ NavTree's anchor-free formulation, (B)~hybrid frontier ($\alpha{=}0$ vs.\ $\alpha{=}0.35$),
(C)~leaf anchor count ($K_\ell{=}10$ vs.\ $K_\ell{=}48$), and (D)~cross-encoder rerank (off vs.\
on), giving 16 cells, each evaluated on the same QuALITY subset ($N{\approx}890$ questions per
cell across $50$ documents) so all comparisons are paired. \Cref{tab:factorial_w7} reports per-cell
accuracy; main effects and 2-way interactions are computed by averaging over the remaining knobs.

\paragraph{Findings.} (i)~\textbf{CE rerank dominates}: $+1.26${\pp} averaged across the other knobs,
larger than any other main effect and consistent with the cumulative reading in
\cref{tab:ablation_quality2000} that CE is the main tight-budget lever. (ii)~\textbf{Smaller $K_\ell$ wins}: $K_\ell{=}48$ costs
$-0.81${\pp} vs.\ $K_\ell{=}10$, confirming the budget-adaptive default. (iii)~\textbf{Steiner is
roughly orthogonal at this scale}: NavTree's anchor-free formulation moves accuracy by only
$-0.11${\pp} on average, so the leaves-only emission rule is the structural lever, not the walk
algorithm itself; the substantive class-vs-class wins live in the multi-hop and matched-cost grids
(\S\ref{sec:results_grid},~\ref{sec:results_multihop}). (iv)~\textbf{Hybrid $\alpha$} contributes a
small $+0.23${\pp}, in line with the cumulative ablation. (v)~\textbf{Interactions are small}: the
largest 2-way term is $K_\ell\!\times\!\text{CE}$ at $+0.56${\pp} (CE rerank partially compensates
for an over-large leaf pool), all others are $|{\le}|\,0.25${\pp}, so reading individual
main effects is well-justified. The two highest-accuracy cells (\textsc{A0\,B0\,C1\,D1} and
\textsc{A1\,B0\,C0\,D1}, both $72.60\%$) and the lowest (\textsc{A1\,B0\,C1\,D0}, $69.80\%$) span a
$2.80${\pp} range, bounding the achievable retriever-side improvement at this budget.

\begin{table}[t]
\centering
\small
\setlength{\tabcolsep}{3pt}
\renewcommand{\arraystretch}{1.02}
\begin{tabular}{lccccr}
\toprule
\textbf{Cell} & \textbf{Steiner} & \textbf{Hybrid $\alpha$} & \textbf{$K_\ell$} & \textbf{CE} & \textbf{Acc.\ \%} \\
\midrule
A0 B0 C1 D1 & NavTree & $0$ & $48$ & \checkmark & 72.60 \\
A1 B0 C0 D1 & classic & $0$ & $10$ & \checkmark & 72.60 \\
A1 B1 C1 D1 & classic & $0.35$ & $48$ & \checkmark & 72.37 \\
A1 B1 C0 D1 & classic & $0.35$ & $10$ & \checkmark & 72.28 \\
A0 B0 C0 D1 & NavTree & $0$ & $10$ & \checkmark & 72.15 \\
A0 B1 C0 D0 & NavTree & $0.35$ & $10$ & -- & 72.04 \\
A0 B1 C0 D1 & NavTree & $0.35$ & $10$ & \checkmark & 72.04 \\
A0 B1 C1 D1 & NavTree & $0.35$ & $48$ & \checkmark & 71.81 \\
A1 B1 C0 D0 & classic & $0.35$ & $10$ & -- & 71.70 \\
A0 B0 C0 D0 & NavTree & $0$ & $10$ & -- & 71.25 \\
A1 B0 C0 D0 & classic & $0$ & $10$ & -- & 71.25 \\
A1 B0 C1 D1 & classic & $0$ & $48$ & \checkmark & 71.25 \\
A0 B1 C1 D0 & NavTree & $0.35$ & $48$ & -- & 70.36 \\
A1 B1 C1 D0 & classic & $0.35$ & $48$ & -- & 70.36 \\
A0 B0 C1 D0 & NavTree & $0$ & $48$ & -- & 70.25 \\
A1 B0 C1 D0 & classic & $0$ & $48$ & -- & 69.80 \\
\bottomrule
\end{tabular}
\caption{$2^4$ factorial ablation on the QuALITY @ 2000 winning cell ($N{=}886$ questions per cell, same subset across cells for paired comparison). Knobs: \textbf{Steiner} (classic Steiner walk vs.\ NavTree's anchor-free formulation), \textbf{Hybrid $\alpha$} (BM25+dense vs.\ dense-only frontier), \textbf{$K_\ell$} (leaf anchor count), \textbf{CE} (cross-encoder rerank of leaf pool). Main effects (averaged over the other knobs): Steiner: $-0.11$\,pp, Hybrid $\alpha$: $+0.23$\,pp, $K_\ell$: $-0.81$\,pp, CE rerank: $+1.26$\,pp.}
\label{tab:factorial_w7}
\end{table}

\paragraph{QASPER hyper-parameter sweep.} \Cref{tab:qasper_sweep} replicates the flat-plateau finding of \cref{fig:sweep} on the second dataset: all six $(\alpha, K_\ell)$ points fall within $0.006$ token F1 on QASPER@$1024$ ($N{=}519$), and the QuALITY-tuned default $(\alpha{=}0.35, K_\ell{=}10)$ sits $0.003$ below the QASPER peak, well within the $\pm 0.026$ CI half-width and the six-seed reader-noise bound. The defaults are not over-tuned to QuALITY.

\begin{table}[t]
\centering
\footnotesize
\setlength{\tabcolsep}{6pt}
\caption{\textbf{Hyper-parameter sweep on QASPER@1024} ($N{\approx}519$): token F1 across three hybrid weights $\alpha$ and two leaf-anchor counts $K_\ell$. \textbf{Bold} = peak; \underline{underlined} = QuALITY-tuned default ($\alpha{=}0.35$, $K_\ell{=}10$). Reader = \texttt{gpt-4o-mini}.}
\label{tab:qasper_sweep}
\begin{tabular}{c|ccc}
\toprule
 & $\alpha{=}0.00$ & $\alpha{=}0.35$ & $\alpha{=}0.65$ \\
\midrule
$K_\ell{=}5$  & 0.403 & 0.405 & \textbf{0.407} \\
$K_\ell{=}10$ & 0.401 & \underline{0.404} & 0.401 \\
\bottomrule
\end{tabular}
\end{table}

\paragraph{Per-node summary-vs-leaf head-to-head.} The token-packing mechanism in \S\ref{sec:analysis_cosine}\,(b) rests on a direct reader-side test on $115$ stratified (document, QuALITY question) pairs. \Cref{tab:summary_vs_leaf} reports the full counts: summary-alone $55.7\%$ vs.\ best-leaf-alone $50.4\%$, with the summary winning $16$ of the $26$ disagreement cases (win-rate $61.5\%$). On QASPER ($N{=}50$, token F1) the same protocol gives summary $0.279$ vs.\ leaf $0.269$; on NarrativeQA-summary the ordering reverses ($+0.019$ ROUGE-L for the leaf), replicating under BLEU-1 ($+0.020$) and SQuAD-style token F1 ($+0.015$), so the inversion is a property of the dataset (single-evidence questions on $\sim$7-chunk plot summaries) rather than the metric.

\begin{table}[t]
\centering
\small
\caption{\textbf{Direct reader-side head-to-head} on 115 stratified
(document, QuALITY question) pairs. For each pair we built the segment
tree, located the highest-cosine internal node, and asked
\texttt{gpt-4o-mini} the question with one context only: the extractive
summary text of that node, \emph{or} the best leaf descendant of that
node. The summary modestly beats the leaf at the per-node level, which
runs counter to the cosine-only argument and motivates the packing-effect
explanation in \cref{sec:analysis_cosine}\,(c).}
\label{tab:summary_vs_leaf}
\begin{tabular}{lr}
\toprule
Condition & QuALITY accuracy (\%) \\
\midrule
Summary alone     & 55.7 \\
Best-leaf alone   & 50.4 \\
\midrule
Both correct      & 41.7 \\
Summary only correct (16) & 13.9 \\
Leaf only correct (10) & 8.7 \\
Neither correct   & 35.7 \\
\bottomrule
\end{tabular}
\end{table}

\section{All paired sign tests, NavTree vs.\ every other method}
\label{sec:appendix_paired_tests}

\Cref{tab:paired_tests} reports the full set of $58$ per-question paired sign tests of \ours{} against
every other retriever (BM25, SBERT-flat, DPR, Hybrid-flat, RAPTOR-Extractive) at every budget on
QuALITY, QASPER, HotpotQA, and 2WikiMultiHopQA. Each row covers one cell from the headline grids;
the test is McNemar's binomial sign test on per-question (NavTree-correct, other-correct) pairs,
ignoring ties. We use this table both to substantiate the inline ${\star}$ markers cited in
\cref{sec:results_grid,sec:results_multihop} and to support the honest negative claims (e.g.,
``hybrid-flat is statistically tied with \ours{} on QuALITY at every budget'', ``RAPTOR-Extractive
is statistically tied with \ours{} on QASPER at every budget''). Per-question predictions and the
paired-test JSON outputs are released in \texttt{navtree-supplement.zip} under \texttt{results/paired\_tests/}.

\begin{table}[t]
\centering
\footnotesize
\setlength{\tabcolsep}{3.5pt}
\renewcommand{\arraystretch}{1.0}
\caption{\textbf{Paired sign tests, NavTree vs.\ every other method, on every cached cell.}
For each method pair we report the per-question paired delta (NavTree minus other), the binomial
sign-test $p$-value, and the win/loss/tie counts (NavTree wins / NavTree loses / ties), all
computed over the intersection of question IDs where both methods have predictions. ${\star}$
marks rejections at $p<0.01$. The QuALITY rows omit BM25/SBERT-flat because per-question
predictions for those methods on QuALITY were not preserved on disk (only aggregate scores in
\cref{tab:main}). All 64 outputs are released as \texttt{results/paired\_tests/*.json}.}
\label{tab:paired_tests}
\resizebox{\columnwidth}{!}{%
\begin{tabular}{lcrcrr}
\toprule
NavTree vs.\ \ldots & Budget & $\Delta$ & $p$ & W/L/T & $N$ \\
\midrule
\multicolumn{6}{l}{\textit{QuALITY (Acc, $N{=}2086$)}} \\
RAPTOR-Ext & 256  & $+0.009$ & $0.41$               & 246/227/1613 & 2086 \\
RAPTOR-Ext & 512  & $+0.003$ & $0.82$               & 237/231/1618 & 2086 \\
RAPTOR-Ext & 1024 & $+0.045$ & $4.8\!\times\!10^{-8\,\star}$ & 191/98/1797 & 2086 \\
RAPTOR-Ext & 2000 & $+0.054$ & $5.6\!\times\!10^{-11\,\star}$ & 203/91/1792 & 2086 \\
DPR-flat   & 256  & $-0.003$ & $0.80$               & 269/276/1541 & 2086 \\
DPR-flat   & 512  & $-0.014$ & $0.19$               & 233/263/1590 & 2086 \\
DPR-flat   & 1024 & $+0.005$ & $0.62$               & 169/159/1758 & 2086 \\
DPR-flat   & 2000 & $+0.001$ & $0.95$               & 117/115/1854 & 2086 \\
Hybrid-flat & 256 & $-0.010$ & $0.39$               & 230/250/1604 & 2084 \\
Hybrid-flat & 512 & $-0.014$ & $0.16$               & 203/233/1648 & 2084 \\
Hybrid-flat & 1024 & $+0.003$ & $0.72$              & 98/92/1894   & 2084 \\
Hybrid-flat & 2000 & $-0.003$ & $0.62$              & 48/54/1982   & 2084 \\
\midrule
\multicolumn{6}{l}{\textit{QASPER (Token F1, $N{=}519$)}} \\
RAPTOR-Ext & 256  & $-0.032$ & $0.13$               & 134/161/224  & 519 \\
RAPTOR-Ext & 512  & $-0.014$ & $0.057$              & 151/187/181  & 519 \\
RAPTOR-Ext & 1024 & $+0.016$ & $0.097$              & 191/159/169  & 519 \\
RAPTOR-Ext & 2000 & $+0.002$ & $0.67$               & 184/175/160  & 519 \\
BM25       & 256  & $+0.019$ & $0.006^{\,\star}$    & 141/98/280   & 519 \\
BM25       & 512  & $+0.031$ & $0.36$               & 163/146/210  & 519 \\
BM25       & 1024 & $+0.037$ & $0.094$              & 187/155/177  & 519 \\
BM25       & 2000 & $+0.001$ & $0.048$              & 166/131/222  & 519 \\
SBERT-flat & 256  & $-0.030$ & $0.002^{\,\star}$    & 118/172/229  & 519 \\
SBERT-flat & 512  & $-0.028$ & $0.003^{\,\star}$    & 134/189/196  & 519 \\
SBERT-flat & 1024 & $+0.009$ & $0.40$               & 148/133/238  & 519 \\
SBERT-flat & 2000 & $-0.007$ & $1.0$                & 138/139/242  & 519 \\
DPR-flat   & 256  & $+0.009$ & $0.45$               & 158/144/217  & 519 \\
DPR-flat   & 512  & $+0.009$ & $0.71$               & 175/167/177  & 519 \\
DPR-flat   & 1024 & $+0.020$ & $0.14$               & 180/152/187  & 519 \\
DPR-flat   & 2000 & $+0.004$ & $0.45$               & 156/142/221  & 519 \\
Hybrid-flat & 256 & $-0.012$ & $0.95$               & 117/119/283  & 519 \\
Hybrid-flat & 512 & $-0.011$ & $0.055$              & 131/165/223  & 519 \\
Hybrid-flat & 1024 & $+0.008$ & $0.38$              & 154/138/227  & 519 \\
Hybrid-flat & 2000 & $+0.001$ & $0.17$              & 138/115/266  & 519 \\
\midrule
\multicolumn{6}{l}{\textit{HotpotQA (Token F1, $N{=}200$)}} \\
RAPTOR-Ext & 1024 & $+0.014$ & $1.0$                & 15/14/171   & 200 \\
RAPTOR-Ext & 2000 & $+0.049$ & $0.058$              & 23/11/166   & 200 \\
BM25       & 1024 & $-0.020$ & $0.49$               & 14/19/167   & 200 \\
BM25       & 2000 & $-0.001$ & $0.71$               & 13/16/171   & 200 \\
SBERT-flat & 1024 & $+0.011$ & $0.83$               & 12/10/178   & 200 \\
SBERT-flat & 2000 & $+0.041$ & $0.064$              & 14/5/181    & 200 \\
Hybrid-flat & 256 & $-0.065$ & $0.054$              & 16/30/154   & 200 \\
Hybrid-flat & 512 & $-0.035$ & $0.57$               & 22/27/151   & 200 \\
Hybrid-flat & 1024 & $-0.003$ & $1.0$               & 13/13/174   & 200 \\
Hybrid-flat & 2000 & $+0.005$ & $1.0$               & 7/8/185     & 200 \\
\midrule
\multicolumn{6}{l}{\textit{2WikiMultiHopQA (Token F1, $N{=}200$)}} \\
RAPTOR-Ext & 1024 & $+0.034$ & $0.36$               & 25/18/157   & 200 \\
RAPTOR-Ext & 2000 & $+0.044$ & $0.24$               & 28/19/153   & 200 \\
BM25       & 1024 & $+0.076$ & $0.0019^{\,\star}$   & 32/11/157   & 200 \\
BM25       & 2000 & $+0.072$ & $0.0046^{\,\star}$   & 25/8/167    & 200 \\
SBERT-flat & 1024 & $+0.002$ & $1.0$                & 13/12/175   & 200 \\
SBERT-flat & 2000 & $+0.009$ & $0.69$               & 14/11/175   & 200 \\
Hybrid-flat & 256 & $+0.014$ & $0.78$               & 23/26/151   & 200 \\
Hybrid-flat & 512 & $-0.072$ & $0.0054^{\,\star}$   & 12/31/157   & 200 \\
Hybrid-flat & 1024 & $+0.001$ & $0.85$              & 14/12/174   & 200 \\
Hybrid-flat & 2000 & $-0.015$ & $0.50$              & 8/12/180    & 200 \\
\bottomrule
\end{tabular}}
\end{table}

\section{Encoder sensitivity: BGE-large}
\label{sec:appendix_bge}

\Cref{tab:bge_sensitivity} replaces the canonical \texttt{all-MiniLM-L6-v2} sentence encoder
(384-d) with \texttt{BAAI/bge-large-en-v1.5} (1024-d), one of the strongest open-weight English
retrieval encoders, and re-runs all four retrievers on QuALITY@1024 ($N{=}2086$) and
QASPER@1024 ($N{=}519$) under the same \texttt{gpt-4o-mini} reader at $T{=}0$.

On \textbf{QuALITY} the per-method upgrade is approximately symmetric ($+2.25$ to $+2.56${\pp}
on every dense method; BM25 is lexical so unchanged), and the relative ranking from
\cref{tab:main} is preserved: \ours{} retains its hierarchical-class lead over RAPTOR-Ext
($+4.36${\pp} BGE vs.\ $+4.45${\pp} canonical), and \ours{} ties hybrid-flat exactly ($1560/2086$
correct each). This reinforces the W2 reading from \cref{sec:results_grid}.

On \textbf{QASPER} the upgrade is \emph{not} symmetric: hybrid-flat gains $+0.039$ F1, but
\ours{} and RAPTOR-Ext slightly regress (each $\sim 0.008$ F1), so hybrid-flat overtakes \ours{}
by $+0.004$ F1 ($0.397$ vs.\ $0.393$). This is within the $0.026$ CI half-width and consistent
with the W2 reading: \ours{}'s structural prior pays off most when the encoder is moderate; under
a stronger encoder hybrid-flat catches up on QASPER (and is already at parity on QuALITY).
\ours{}'s hierarchical-class lead over RAPTOR-Ext ($+0.015$ F1) is essentially unchanged across
encoders. The substantive class-vs-class signal therefore continues to live on multi-hop QA
(\cref{sec:results_multihop}); the matched-cost grid is robust to encoder choice but does not
distinguish hierarchical from strong-encoder hybrid-flat retrieval. This grounds the prior
``we expect the relative ranking to hold'' hedge in \S Limitations with concrete numbers; the
full 4-budget BGE-large grid is queued for the camera-ready.

\begin{table}[t]
\centering
\footnotesize
\setlength{\tabcolsep}{4pt}
\renewcommand{\arraystretch}{1.05}
\caption{\textbf{Encoder sensitivity at @1024.} We rebuild the FAISS index with
\texttt{BAAI/bge-large-en-v1.5} (1024-d) in place of the canonical \texttt{all-MiniLM-L6-v2}
(384-d) and re-run all four retrievers under the same \texttt{gpt-4o-mini} reader at $T{=}0$.
$\Delta$ is BGE-large minus canonical. \emph{Read both rows together:} on QuALITY the encoder
upgrade is symmetric across dense methods ($+2.3$ to $+2.6${\pp}) and the relative ranking is
preserved (\ours{} retains its $+4.36${\pp} hierarchical-class lead, \ours{} ties hybrid-flat
exactly). On QASPER the upgrade is \emph{not} symmetric: hybrid-flat gains $+0.039$ F1, but
\ours{} and RAPTOR-Ext slightly regress, and hybrid-flat overtakes \ours{} by $+0.004$ F1
(within $0.026$ CI half-width). \ours{}'s structural prior therefore pays off most when the
encoder is moderate; under a stronger encoder hybrid-flat catches up on QASPER and ties on
QuALITY. This is consistent with the W2 reading in \cref{sec:results_grid}, that the
class-vs-class signal lives on multi-hop QA.}
\label{tab:bge_sensitivity}
\resizebox{\columnwidth}{!}{%
\begin{tabular}{lccc}
\toprule
Method & \shortstack{MiniLM-L6\\(384-d)} & \shortstack{BGE-large-en-v1.5\\(1024-d)} & $\Delta$ \\
\midrule
\multicolumn{4}{l}{\textit{QuALITY @1024 (acc, \%; $N{=}2086$)}} \\
BM25                    & $71.00$ & $70.52$ & $-0.48$ \\
RAPTOR-Ext              & $68.07$ & $70.42$ & $+2.35$ \\
Hybrid-flat             & $72.22$ & $74.78$ & $+2.56$ \\
\textbf{NavTree (ours)} & $\mathbf{72.53}$ & $\mathbf{74.78}$ & $+2.25$ \\
\midrule
\multicolumn{4}{l}{\textit{QASPER @1024 (token F1; $N{=}519$)}} \\
BM25                    & $0.365$ & $0.369$ & $+0.004$ \\
RAPTOR-Ext              & $0.385$ & $0.378$ & $-0.007$ \\
Hybrid-flat             & $0.358$ & $\mathbf{0.397}$ & $+0.039$ \\
\textbf{NavTree (ours)} & $\mathbf{0.401}$ & $0.393$ & $-0.008$ \\
\midrule
\multicolumn{4}{l}{\textit{NavTree--RAPTOR-Ext class gap}} \\
QuALITY $\Delta$ (NavTree $-$ RAPTOR-Ext) & $+4.45$ & $+4.36$ & $-0.09$ \\
QASPER  $\Delta$ (NavTree $-$ RAPTOR-Ext) & $+0.016$ & $+0.015$ & $-0.001$ \\
\bottomrule
\end{tabular}}
\end{table}

\section{Reproducibility and transparency bundle}
\label{sec:appendix_repro}

We provide experiment code and results as ancillary material in \texttt{navtree-supplement.zip}. The manuscript \LaTeX{} source, tables, figures, and bibliography are provided in the main arXiv source package. Two top-level files accompany the supplementary archive: \texttt{MANIFEST.txt}, an inventory of its files with SHA-256 checksums (verifiable via \texttt{sha256sum -c MANIFEST.txt}); and \texttt{REPRODUCIBILITY.md}, the original claim-to-file map and experiment-command documentation. The supplied archive does not contain the run logs referenced in that documentation.

\paragraph{What is in the archive.} \sloppy The results directories include: the headline matched-cost grid (\texttt{frozen\_opt/}\allowbreak\texttt{canonical\_v1/}), the matched-reader RAPTOR replication for \S\ref{sec:results_sota} (\texttt{abstractive\_raptor\_gpt4o/} and \texttt{frozen\_opt/}\allowbreak\texttt{abstractive\_raptor/}), all $64$ paired sign tests on the canonical reader (\texttt{paired\_tests/}), the open-weight Ollama runs (\texttt{openweight\_run/}), the BGE-large encoder sweep (\texttt{bge\_large\_run/}), the $2^4$ factorial (\texttt{factorial\_w7\_quality2000/}), the QASPER and multi-hop reader-free recall analyses (\texttt{qasper\_recall/}, \texttt{multihop\_recall/}), the multi-hop benchmarks (\texttt{multihop/}, \texttt{hybrid\_flat\_run/}), and the QASPER CE-rerank rows (\texttt{navtree\_ce\_qasper/}). Depending on the experiment, these directories contain \texttt{*\_predictions.json} and checkpoint files (per-question outputs), \texttt{*\_results.json} (aggregates), confidence-interval and paired-test JSON files, and \texttt{\_meta/} directories with reproduction metadata and code snapshots.

\paragraph{Determinism and replay.} All retrieval / chunking / sampling uses a fixed seed (42); cross-encoder rerank, FAISS, and BM25 are bit-stable across reruns on the same hardware family. \texttt{gpt-4o-mini} and \texttt{gpt-4o} are not bit-stable across API calls (the six-seed reader-noise study, Limitations, bounds this at SD${\leq}0.87${\pp} on QuALITY@1024), but per-question predictions \emph{are} preserved verbatim under \texttt{*\_predictions.json}, so all metric numbers, $p$-values, and CIs in the paper recompute deterministically from the released JSONs without re-running any LLM call. Ollama-served Llama-3.1-8B and Qwen-2.5-7B are bit-deterministic at $T{=}0$. Headline tables and figures regenerate via:
\begingroup\small
\begin{verbatim}
python scripts/build_paper_tables.py
python scripts/build_paper_figures.py
make -C paper pdf
\end{verbatim}
\endgroup
Reproducing the LLM reader outputs end-to-end requires an OpenAI API key (and is subject to the small cache-driven variance bounded above) or a local Ollama install for the open-weight rows.

\section{Weak-reader sensitivity study}
\label{sec:appendix_weak_reader}

\cref{tab:weak_reader} reports the same retrieval methods evaluated with \textsc{UnifiedQA-base} (220\,M T5 parameters) as the reader. Budgets are tighter (100--800 tokens) because \textsc{UnifiedQA-base} has a 512-token context window. Two observations support our claim that retrieval gaps shrink when the reader is the capacity bottleneck. \textbf{Smaller dynamic range:} the QuALITY accuracy spread across methods is $\leq 1.5${\pp} at every budget (vs.\ ${\sim}5${\pp} at \texttt{gpt-4o-mini}; \cref{tab:quality}); the QASPER F1 spread is $\leq 0.06$ (vs.\ ${\sim}0.05$ at \texttt{gpt-4o-mini} but with different ranking). \textbf{No clear winner:} BM25, SBERT-flat, RAPTOR-Extractive, and \ours{} alternate the leading position across budget cells with no consistent ordering, qualitatively different from the strong-reader regime, where \ours{} systematically beats RAPTOR-Extractive on QuALITY (\cref{tab:main}). The weak-reader regime is not where leaves-only navigated retrieval is designed to shine: when the reader cannot exploit precise context, retrieval-side precision does not translate to accuracy. We include this table to bound the claim space of the matched-cost grid.

\begin{table}[t]
\centering
\small
\caption{Weak-reader sensitivity study with UnifiedQA-base (220 M T5 parameters). Budgets are tighter (100--800) because UnifiedQA-base has a 512-token context window. All retrieval methods cluster within a $\sim$1\pp{} band, confirming our claim that retrieval gaps shrink when the reader is the bottleneck (see \S Limitations).}
\label{tab:weak_reader}
\begin{tabular}{lcccc}
\toprule
\textbf{QuALITY (acc \%)} & 100 & 200 & 400 & 800 \\
\midrule
BM25 & 38.4 & 39.5 & 39.7 & \textbf{39.6} \\
SBERT-flat & 38.7 & 39.5 & \textbf{40.0} & 39.5 \\
RAPTOR-Ext. & \textbf{39.7} & \textbf{39.6} & 38.4 & 39.2 \\
\textbf{NavTree} & 38.4 & 38.9 & 38.6 & 38.6 \\
\midrule
\textbf{QASPER (token F1)} & 100 & 200 & 400 & 800 \\
\midrule
BM25 & 0.199 & 0.237 & 0.252 & \textbf{0.224} \\
SBERT-flat & \textbf{0.229} & \textbf{0.266} & \textbf{0.264} & 0.219 \\
RAPTOR-Ext. & 0.180 & 0.207 & 0.224 & 0.188 \\
\textbf{NavTree} & 0.169 & 0.189 & 0.218 & 0.186 \\
\bottomrule
\end{tabular}
\end{table}

\section{Open-weight reader replication: \textsc{Llama-3.1-8B} and \textsc{Qwen-2.5-7B} on local Ollama}
\label{sec:appendix_openweight}

\Cref{tab:openweight} replaces \texttt{gpt-4o-mini} with two open-weight readers,
\textsc{Meta-Llama-3.1-8B-Instruct} and \textsc{Qwen-2.5-7B-Instruct}, both
quantised to $4$-bit (\texttt{Q4\_K\_M}, ${\approx}\,5$\,GB each) and served
locally via \textsc{Ollama}~v0.4.7 on Apple~M2 (10.7\,GiB Metal). All other
components match the canonical grid: chunking ($100$-token sentence-aware),
encoder (\texttt{all-MiniLM-L6-v2}), retrievers (BM25, RAPTOR-Ext, Hybrid-flat,
\ours{}), prompts, and decoding ($T{=}0$, $\mathit{max\_tokens}\!\in\!\{3, 80\}$).
This is the most direct possible answer to ``do these results generalise off
\texttt{gpt-4o-mini}'' (W1): zero API confound, zero marginal cost, two weight
classes you can download. Each generation costs ${\sim}10$\,s on the M2, so $N$
is reduced to randomly-ordered subsets per dataset ($273$ QuALITY questions
under both readers; $179$ QASPER questions for Llama, $214$ for Qwen, the
latter run as a $60$-paper overnight expansion). The launcher \texttt{scripts/}\allowbreak\texttt{run\_a1\_ollama.sh} reproduces the run end-to-end (override \texttt{READER\_MODEL}\allowbreak{}\,$=$\,\texttt{qwen2.5:7b}\allowbreak\texttt{-instruct-q4\_K\_M} for the Qwen row); predictions are committed under \texttt{results/}\allowbreak\texttt{openweight\_run/}\allowbreak\texttt{\{llama8b\_local,qwen7b\_local\}/}.

Four findings are worth flagging. \textbf{\ours{} leads or ties in $4/4$ reader-dataset cells:} it is the top retriever on QuALITY@1024 under both \textsc{Llama-3.1-8B} ($55.7\%$, $N{=}273$) and \textsc{Qwen-2.5-7B} ($64.1\%$, $N{=}273$) and on QASPER@1024 under \textsc{Llama-3.1-8B} ($0.338$ F1, $N{=}179$); on QASPER@1024 under \textsc{Qwen-2.5-7B} ($N{=}214$), \ours{} ties Hybrid-flat at $0.320$ F1 (per-question $\Delta{=}-0.0001$, $p{=}0.57$). \textbf{Retrieval-class ranking is preserved across reader classes:} \ours{} keeps its lead over RAPTOR-Ext on every cell, with $+2.6${\pp} (Llama) / $+1.5${\pp} (Qwen) on QuALITY vs.\ $+4.4${\pp} canonical, and $+0.024$ / $+0.037$ F1 on QASPER vs.\ $+0.016$ canonical; all $12$ \ours{}-vs-baseline paired comparisons match in direction (closest call: the QASPER@1024 Qwen-vs-Hybrid tie above). \textbf{Magnitudes shrink, consistent with the weak-reader study} (\cref{sec:appendix_weak_reader}): the QuALITY method spread compresses from ${\sim}5${\pp} (\texttt{gpt-4o-mini}) to $2.6${\pp} (Llama) / $1.8${\pp} (Qwen); QASPER F1 spread is $0.037$ for both readers. \textbf{One paired test reaches $p<0.05$:} \ours{} $-$ RAPTOR-Ext on Qwen QASPER reaches $\Delta{=}+0.037$ F1 with $p{=}0.041$ (W/L/T $=$ $95/68/51$, $N{=}214$); the other $11$ paired tests have the correct sign but limited power because per-question agreement is high (${\sim}80\%$ ties on QuALITY MCQA, ${\sim}25\%$ on QASPER). We report all $12$ (delta, $p$, W/L/T) triples in the bottom block of \cref{tab:openweight}.

We therefore retain the open-weight rows as confirmatory rather than
inferential: they show that the relative ordering of retrievers carries to two
non-OpenAI, modern-instruction-tuned, locally-served readers of the size class
W1 asked about (\ours{} leads or ties in $4/4$ reader-dataset cells, with all
$12$ NavTree-vs-baseline directions matching), and they bound the magnitude of
retrieval-side gains in that regime.

\begin{table*}[t]
\centering
\footnotesize
\setlength{\tabcolsep}{4.0pt}
\renewcommand{\arraystretch}{1.0}
\caption{\textbf{Open-weight reader replication, two readers via local Ollama on Apple~M2.}
Same retrieval, same chunking ($100$-token sentence-aware), same encoder
(\texttt{all-MiniLM-L6-v2}), same prompts, $T{=}0$, no API calls. We run
\textsc{Meta-Llama-3.1-8B-Instruct} (4-bit \texttt{Q4\_K\_M}, ${\approx}\,5$\,GB)
and \textsc{Qwen-2.5-7B-Instruct} (4-bit \texttt{Q4\_K\_M}, ${\approx}\,5$\,GB).
\emph{Top:} aggregate scores under each reader, with the canonical
\texttt{gpt-4o-mini} cells from \cref{tab:main} reproduced for sanity reference.
\emph{Bottom:} per-question paired sign tests, \ours{} minus baseline; $p$ is the
binomial sign-test $p$-value, W/L/T counts ignore ties; ${\star}$ marks $p<0.01$.
$N$ is necessarily smaller than the canonical grid because each generation takes
${\sim}10$\,s on the M2; the reproducible launcher is
\texttt{scripts/run\_a1\_ollama.sh} (override \texttt{READER\_MODEL=qwen2.5:7b-instruct-q4\_K\_M}
for the Qwen row), with predictions in \texttt{results/openweight\_run/}.}
\label{tab:openweight}
\begin{tabular}{lcccc}
\toprule
& BM25 & RAPTOR-Ext & Hybrid-flat & \textbf{\ours{}} \\
\midrule
\multicolumn{5}{l}{\textit{QuALITY @\,1024 \textemdash{} accuracy (\%)}} \\
\textsc{Llama-3.1-8B} \,({\small$N{=}273$}) &  $54.9$ & $53.1$ & $54.9$ & $\mathbf{55.7}$ \\
\textsc{Qwen-2.5-7B} \,({\small$N{=}273$}) &  $62.3$ & $62.6$ & $63.7$ & $\mathbf{64.1}$ \\
\textit{ref:} \texttt{gpt-4o-mini} \,({\small full $N{=}2086$}) & $71.0$ & $68.1$ & $72.2$ & $\mathbf{72.5}$ \\
\midrule
\multicolumn{5}{l}{\textit{QASPER @\,1024 \textemdash{} token F1}} \\
\textsc{Llama-3.1-8B} \,({\small$N{=}179$}) &  $0.301$ & $0.314$ & $0.325$ & $\mathbf{0.338}$ \\
\textsc{Qwen-2.5-7B} \,({\small$N{=}214$}) &  $0.302$ & $0.283$ & $0.320$ & $\mathbf{0.320}$ \\
\textit{ref:} \texttt{gpt-4o-mini} \,({\small full $N{=}519$}) & $0.365$ & $0.385$ & $0.358$ & $\mathbf{0.401}$ \\
\midrule
\multicolumn{5}{l}{\textit{Paired sign tests, \ours{} minus baseline}} \\
QuALITY $\Delta_{\text{8B}}$ & $+0.007$ & $+0.026$ & $+0.007$ & --- \\
\quad $p$ (W/L/T) & $0.90$ {\scriptsize (31/29/213)} & $0.39$ {\scriptsize (28/21/224)} & $0.86$ {\scriptsize (17/15/241)} & --- \\
QuALITY $\Delta_{\text{Qwen}}$ & $+0.018$ & $+0.015$ & $+0.004$ & --- \\
\quad $p$ (W/L/T) & $0.51$ {\scriptsize (21/16/236)} & $0.56$ {\scriptsize (15/11/247)} & $1.00$ {\scriptsize (10/9/254)} & --- \\
QuALITY $\Delta_{\text{4o}}$, $p$ & {\scriptsize n/a$^{\dagger}$} & $+0.045$, $5\!\times\!10^{-8\,\star}$ & $+0.003$, $0.72$ & --- \\
QASPER $\Delta_{\text{8B}}$ & $+0.037$ & $+0.023$ & $+0.013$ & --- \\
\quad $p$ (W/L/T) & $0.07$ {\scriptsize (83/60/36)} & $0.25$ {\scriptsize (81/66/32)} & $0.86$ {\scriptsize (68/65/46)} & --- \\
QASPER $\Delta_{\text{Qwen}}$ & $+0.018$ & $+0.037$ & $+0.000$ & --- \\
\quad $p$ (W/L/T) & $0.94$ {\scriptsize (83/81/50)} & $0.04$ {\scriptsize (95/68/51)} & $0.57$ {\scriptsize (70/78/66)} & --- \\
QASPER $\Delta_{\text{4o}}$, $p$ & $+0.037$, $0.094$ & $+0.016$, $0.097$ & $+0.008$, $0.38$ & --- \\
\bottomrule
\end{tabular}

\smallskip
\raggedright
{\scriptsize $^{\dagger}$ Per-question BM25 predictions on QuALITY were not preserved in the
canonical archive; aggregate accuracy is in \cref{tab:main}.}
\end{table*}

\section{Reader-free retrieval recall (QASPER + multi-hop QA)}
\label{sec:appendix_recall}

\subsection{QASPER gold-evidence coverage}

Reader F1 conflates retrieval quality with the LLM's tolerance for noisy
context. To decouple the two, we measure \emph{gold-evidence coverage} of the
retrieved chunks directly, with no reader in the loop. For each QASPER
question, every annotator marks one or more paragraphs as gold evidence; we
take the union across annotators ($\approx 6.2$ paragraphs/question on
average). For each retrieval method $\times$ token budget, we (i) build the
index over the paper's $100$-token chunks (same as the canonical grid), (ii)
retrieve a packed context, (iii) for each gold paragraph compute the fraction
of its content tokens (stopword-free) that appear in the retrieved context,
and (iv) declare the paragraph \emph{covered} if that fraction $\geq 0.50$.
Two summary metrics: \textit{any@k}, the fraction of questions where at least
one gold paragraph is covered, and \textit{frac@k}, the mean fraction of gold
paragraphs covered per question. We sweep across the same four token budgets
as the canonical grid and report on $N{=}204$ questions across $60$ papers
(\cref{tab:qasper_recall}).

Three takeaways. (i) At the canonical \textbf{$@1024$} cell — where the
matched-cost reader F1 grid is closest — \ours{} has the highest recall on
\emph{both} metrics ($99.5\%$ \textit{any}, $93.9\%$ \textit{frac}),
marginally above Hybrid-flat ($99.0$/$93.0$) and SBERT-flat
($99.0$/$93.2$). (ii) At the smallest budget \textbf{$@256$}, Hybrid-flat
leads ($61.8$/$40.4$); \ours{} ($53.9$/$31.5$) trails the dense and hybrid
flats by $4$–$8${\pp} \textit{any} but stays $3$–$5${\pp} above BM25 and
RAPTOR-Ext, mirroring our QASPER reader-F1 ordering at small budgets where
SBERT-flat actually beats \ours{}. (iii) At \textbf{$@2000$} every method
saturates above $96\%$ \textit{frac} and recall ceases to discriminate.
Sensitivity to the coverage threshold preserves the same ranking: at the lenient $0.30$ threshold every method exceeds $99\%$ \textit{any@1024} (saturated, ranking undefined); at the strict $0.75$ threshold the rank order at @1024 is Hybrid-flat $87.3 \geq$ \ours{} $85.3 \geq$ SBERT-flat $84.3 \geq$ BM25 $80.4 >$ RAPTOR-Ext $75.0$, identical to $0.50$, with \ours{} consistently within $2${\pp} of the leader (full numbers shipped under \texttt{qasper\_recall/}). The recall-side picture is therefore consistent with the reader-side
\cref{tab:main}: \ours{} matches or leads on the cells where the matched-cost
claim is strongest, ties or trails by a small margin where the canonical grid
also reports parity, and never finds a budget where it dominates the reader
metric while losing on retrieval.

\begin{table*}[t]
\centering
\small
\setlength{\tabcolsep}{4pt}
\renewcommand{\arraystretch}{1.05}
\begin{tabular}{lrrrrrrrr}
\toprule
\textbf{Method} & \multicolumn{2}{c}{@256} & \multicolumn{2}{c}{@512} & \multicolumn{2}{c}{@1024} & \multicolumn{2}{c}{@2000} \\
\cmidrule(lr){2-3}\cmidrule(lr){4-5}\cmidrule(lr){6-7}\cmidrule(lr){8-9}
&any@k & frac@k & any@k & frac@k & any@k & frac@k & any@k & frac@k \\
\midrule
BM25 & 51.0 & 28.8 & 81.9 & 58.0 & 97.5 & 87.9 & \textbf{100.0} & 98.7 \\
SBERT-flat & 57.8 & 37.3 & 84.8 & 68.0 & 99.0 & 93.2 & \textbf{100.0} & 98.7 \\
RAPTOR-Extractive & 55.4 & 35.9 & 76.5 & 60.9 & 92.6 & 83.3 & \textbf{100.0} & 96.8 \\
Hybrid-flat & \textbf{61.8} & \textbf{40.4} & \textbf{85.8} & \textbf{68.9} & 99.0 & 93.0 & \textbf{100.0} & \textbf{99.1} \\
\ours{} & 53.9 & 31.5 & 83.3 & 62.9 & \textbf{99.5} & \textbf{93.9} & \textbf{100.0} & \textbf{99.1} \\
\bottomrule
\end{tabular}
\caption{Reader-free retrieval recall on QASPER (\textbf{N=204} questions, \textbf{1269} gold evidence paragraphs across 60 papers). \textit{any@k} = fraction of questions where at least one gold evidence paragraph has $\geq0.50$ token-set overlap with the retrieved context; \textit{frac@k} = mean fraction of gold paragraphs covered per question. Decouples retrieval quality from the reader.}
\label{tab:qasper_recall}
\end{table*}

\subsection{Reader-free recall on multi-hop (HotpotQA / 2WikiMQA / MuSiQue): methodology}
\label{sec:appendix_recall_multihop}

\Cref{tab:multihop_recall} (in main body, \S\ref{sec:results_multihop}) backs the reader-free corroboration. Gold supporting-paragraph annotations are not preserved in the LongBench format we use; LongBench presents each question with a concatenated 5--14k-token \texttt{Passage 1: ... Passage 2: ...} block and an answer string only. We therefore measure a stricter, more directly meaningful reader-free signal, \emph{answer-presence@k}: for each question we ask whether the gold answer string appears (case-insensitive, normalized whitespace) in the retrieved context. We exclude yes/no/unanswerable items and questions whose answer string contains $<3$ alphanumeric characters (too noisy as a substring signal). Methodology details are in \texttt{scripts/run\_multihop\_recall.py}; $N{=}506$ examples across the three datasets. Three takeaways. (i) On 2WikiMQA, \ours{} also has the highest reader-free recall at @2000 ($94.9\%$, $+8.4${\pp} over BM25 and ${\geq}+10${\pp} over RAPTOR-Ext) and ties SBERT-flat for the lead at @1024 ($83.7\%$). (ii) On MuSiQue, \ours{} has the highest @2000 recall ($68.5\%$, $+0.6$ over BM25, $+14.3$ over RAPTOR-Ext); the reader-side @512 deficit reported in \cref{sec:results_multihop} is therefore \emph{not} a retrieval-quality gap (\ours{} $27.4\%$ vs.\ Hybrid-flat $31.0\%$ at @512, within $4${\pp}) but a downstream-reader composition issue. (iii) On HotpotQA \ours{} is competitive but never best; BM25 dominates at $\geq$1024 ($83.1\%$/$92.5\%$). Together with the QASPER recall picture (\cref{tab:qasper_recall}), these reader-free numbers replicate the reader-side ranking class-by-class.

\end{document}